\documentclass[runningheads]{llncs}

\usepackage{eccv}

\usepackage{eccvabbrv}

\usepackage{graphicx}
\usepackage{booktabs}

\usepackage[accsupp]{axessibility}  % Improves PDF readability for those with disabilities.
\usepackage{multirow}  % Add this package for multirow support

\usepackage{hyperref}

\usepackage{orcidlink}

\begin{document}

\title{MobileManiBench: Simplifying Model
Verification for Mobile Manipulation\vspace{-2mm}} 

% \titlerunning{MobileManiBench}

\author{Wenbo Wang$^{2}$ \quad Fangyun Wei$^{1}$ \quad Qixiu Li$^{3}$ \quad Xi Chen$^{1}$ \quad Yaobo Liang$^{1}$ \quad Chang Xu$^{2}$\thanks{Corresponding author.} \quad Jiaolong Yang$^{1}$ \quad Baining Guo$^{1}$
\vspace{-2mm}
}

% % TODO FINAL: Replace with an abbreviated list of authors.
\authorrunning{Wenbo Wang et al.}

% % TODO FINAL: Replace with your institution list.
% Springer Heidelberg, Tiergartenstr.~17, 69121 Heidelberg, Germany

\institute{$^1$Microsoft Research Asia \quad $^2$University of Sydney \quad $^3$Tsinghua University \\
\texttt{\small\{fawe,xichen6,yalia,jiaoyan,bainguo\}@microsoft.com}\\
\texttt{\small wwan0412@uni.sydney.edu.au} \quad \texttt{\small liqx23@mails.tsinghua.edu.cn}  \quad \texttt{\small c.xu@sydney.edu.au}}

\maketitle

\begin{abstract}
\vspace{-6mm}
  Vision-language-action models have advanced robotic manipulation but remain constrained by reliance on the large, teleoperation-collected datasets dominated by the static, tabletop scenes. We propose a simulation-first framework to verify VLA architectures before real-world deployment and introduce MobileManiBench, a large-scale benchmark for mobile-based robotic manipulation. Built on NVIDIA Isaac Sim and powered by reinforcement learning, our pipeline autonomously generates diverse manipulation trajectories with rich annotations (language instructions, multi-view RGB–depth–segmentation images, synchronized object/robot states and actions). MobileManiBench features 2 mobile platforms (parallel-gripper and dexterous-hand robots), 2 synchronized cameras (head and right wrist), 630 objects in 20 categories, 5 skills (open, close, pull, push, pick) with over 100 tasks performed in 100 realistic scenes, yielding 300K trajectories. This design enables controlled, scalable studies of robot embodiments, sensing modalities, and policy architectures, accelerating research on data efficiency and generalization. We benchmark representative VLA models and report insights into perception, reasoning, and control in complex simulated environments, with all code, datasets, and models publicly released at our project website: \href{https://dexhand.github.io/MobileManiBench/}{https://dexhand.github.io/MobileManiBench/}.
  \vspace{-2mm}
  \vspace{-8mm}

\end{abstract}

\section{Introduction}
\label{sec:intro}
\vspace{-2mm}

Recent advances in Vision-Language-Action models~\cite{team2024octo,o2024open,li2024cogact,qu2025spatialvla,black2410pi0,intelligence2025pi_,kim2025fine,brohan2022rt,li2022vision,zitkovich2023rt,liu2024rdt,wen2025dexvla,bjorck2025gr00t,wang2025unified, zhong2025dexgraspvla, cheang2025gr, li2025scalable,team2025gemini,yang2025magma,shi2025memoryvla,zhang2025dreamvla} have substantially advanced robotic manipulation, demonstrating strong generalization to novel objects and diverse visual domains. However, the success of current VLA models heavily dependents on large-scale datasets, among which, the \textit{openX-Embodiment} dataset~\cite{o2024open} has become the de facto training resource. Yet, most of its data are collected via teleoperation from static, third-person or head-mounted viewpoints, restricted to indoor tabletop environments populated with a limited set of household objects. These constraints introduce several challenges:
\begin{itemize}
\vspace{-2mm}
    \item Any modification to the hardware configuration—such as incorporating new sensors (e.g., depth or wrist-mounted cameras), extending to mobile-based manipulation, or replacing grippers with dexterous hands—necessitates recollecting data from scratch. If these additions fail to provide substantial performance gains, the effort becomes wasteful due to the high cost and inefficiency of teleoperation-based data collection.
    \item Although teleoperation pipelines are effective for simple gripper-based robots, they become cumbersome when scaled to dexterous or mobile platforms, thereby impeding rapid model development and iteration.
\end{itemize}

\begin{figure}[!t]
    \centering
    \includegraphics[width=0.99\linewidth]{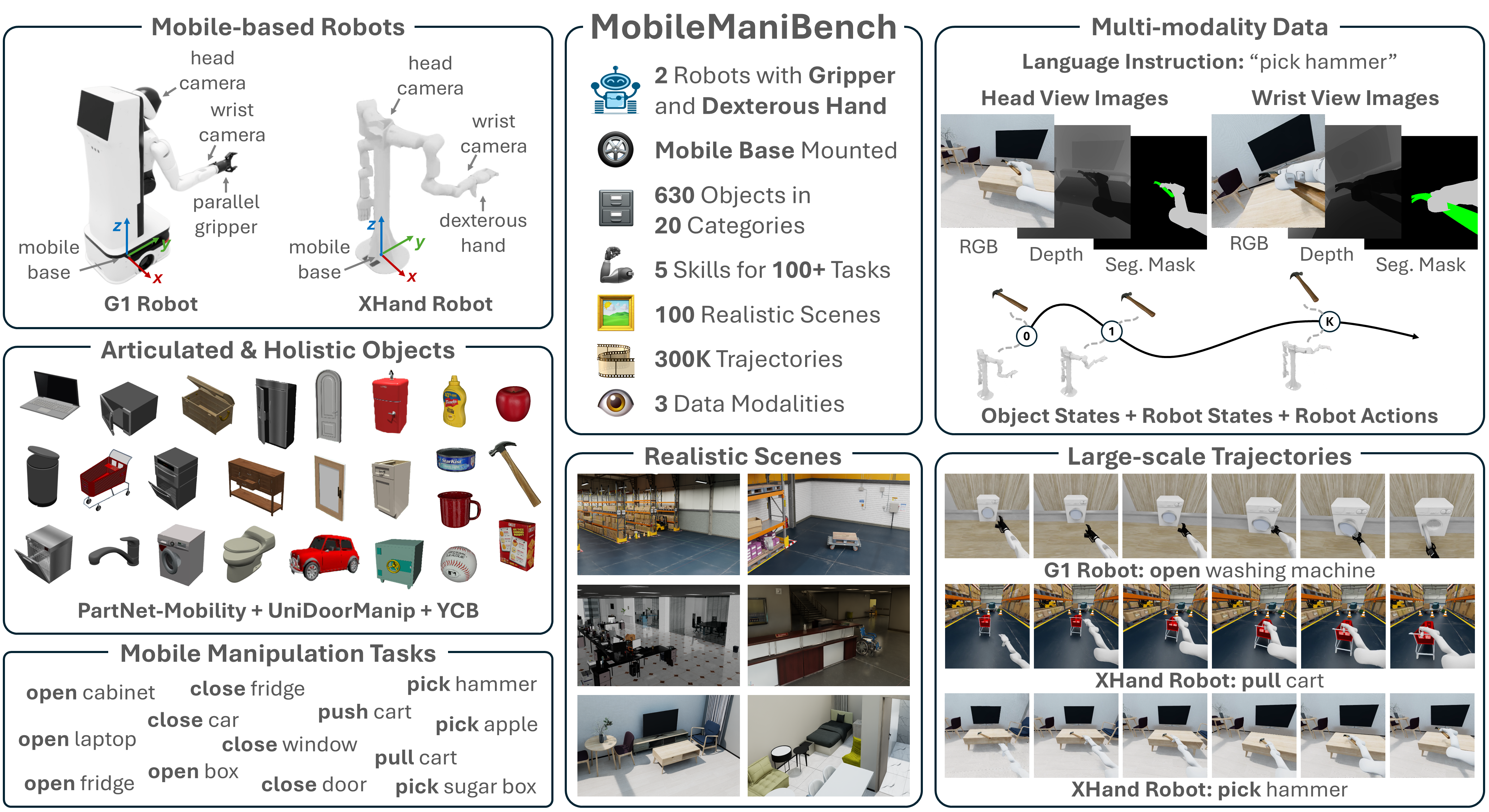}
    \vspace{-2mm}
    \caption{
    Overview of \textbf{MobileManiBench}. It features 2 mobile-based robots: the G1 robot with a parallel gripper and the XHand robot with a dexterous hand. The benchmark includes 630 articulated and holistic objects across 20 categories and supports 5 mobile manipulation skills—open, close, pull, push, and pick—enabling over 100 tasks. To efficiently scale data generation while ensuring task success, we train a universal \textbf{MobileManiRL} policy for each robot–object–skill triplet and generate \textbf{MobileManiDataset} across 100 realistic scenes with 300K trajectories and 3 data modalities—language instructions, multi-view RGB–depth–segmentation images, synchronized object/robot states and actions. MobileManiBench offers a flexible testbed to accelerate model innovation and data-efficiency research for VLA models.
    }
    \vspace{-6mm}
    \label{fig:teaser}
\end{figure}

To support scalable \textit{verification} of VLA architectures, we propose conducting data generation and evaluation in simulation environments prior to collecting real-world robotic data. The key idea is to offer an efficient way to scale up objects, tasks, and scenes, while flexibly supporting diverse robot configurations—such as dexterous hands, mobile bases, arbitrary camera placements—and easily acquiring multi-modal sensory inputs (e.g., RGB, depth, robot proprioception). In this work, we leverage \textit{NVIDIA Isaac Sim}~\cite{NVIDIA_Isaac_Sim} as the simulation platform and employ reinforcement learning~\cite{PPO} to train agents that can automatically generate manipulation trajectories along with corresponding annotations (e.g., text instructions, camera images, and robot states). Building upon this automated pipeline, we introduce \textbf{MobileManiBench}—a large-scale benchmark for mobile manipulation and embodied skill learning.

As shown in Figure~\ref{fig:teaser}, our MobileManiBench is designed with several distinguishing features:
\begin{itemize}
    \item \textit{Multi-Robot Embodiment}: Two robot platforms are provided—one equipped with a parallel gripper and another with a dexterous anthropomorphic hand.
    \item \textit{Mobile-Base Platform}: Both robots are mounted with a mobile base, allowing for spatially extended tasks that require coordinated navigation and manipulation.
    \item \textit{Rich Visual Sensing}: Two cameras are mounted on the head and right wrist, each capturing synchronized RGB, depth, and segmentation images to facilitate multi-view and multi-sensory perception.
    \item \textit{Multi-Modality Trajectories}: Each recorded trajectory includes textual instructions, multi-view visual images, together with object states, robot states and actions, forming  comprehensive supervision signals for VLA training.
    \item \textit{Diverse and Scalable Content}: The dataset contains 630 unique objects spanning 20 categories, 100 realistic and semi-structured scenes, 5 mobile manipulation skills across 100 tasks, and generates over 300K trajectories.
\end{itemize}

The above design enables controlled experimentation with different robot configurations, sensory modalities, and model architectures, without requiring expensive real-world data collection. Researchers can efficiently evaluate whether new input modalities (e.g., wrist or depth images) contribute to better manipulation performance or whether specific architectural designs improve multi-sensory fusion and reasoning. MobileManiBench provides a flexible testbed for accelerating both model innovation and data-efficiency exploration in embodied AI.

Leveraging MobileManiBench, we systematically compare several existing VLA models~\cite{black2410pi0,intelligence2025pi_,kim2024openvla,li2024cogact} with our proposed MobileManiVLA, analyze their strengths and weaknesses across various tasks, and provide a series of empirical takeaways that offer new insights into how large-scale multi-modal architectures perceive and act in complex robotic manipulation environments. We hope that this benchmark will foster reproducible research, enable rapid architectural iteration, and pave the way toward scalable, general-purpose VLA models.

\vspace{-4mm}
\section{Related Works}
\label{sec:related-works}
\vspace{-2mm}

\begin{table*}[!t]
\centering
\resizebox{0.99\textwidth}{!}{\begin{tabular}{l|cccc|ccccc}
\toprule
\multirow{2}{*}{Benchmarks / Datasets} & Object & Skill & Scene & Trajectory 
& Mobile & Parallel & Dexterous & Articulated & Universal \\
 & Number & Number$^{\dagger}$ & Number & Number & Based & Gripper & Hand & Object & Model$^{\dagger\dagger}$ \\

\midrule
% real-world
DROID~\cite{khazatsky2025droid}
& * & 86 & 564 & 76K &  & \checkmark &  & \checkmark &  \\
Open X‑Embodiment~\cite{o2024open}
& 5228 & 527 & 311 & 1400K &  & \checkmark &  & \checkmark &  \checkmark \\
AgiBotWorld-Beta~\cite{agibotworldcontributors2025agibotworld}
& 3000 & 87 & 106 & 1000K &  & \checkmark & \checkmark & \checkmark & \checkmark \\
DexGraspVLA~\cite{zhong2025dexgraspvla}
& 36 & 1 & - & 2K &  &  & \checkmark &  & \checkmark \\
Humanoid Everyday~\cite{zhao2025humanoideveryday}
& * & 221 & * & 10.3K & \checkmark &  & \checkmark & \checkmark & \checkmark \\
Mobile-ALOHA~\cite{fu2024mobilealoha}
& 6 & 6 & 6 & 300 & \checkmark & \checkmark &  & \checkmark &  \\
\midrule
% simulation
% UniGraspTransformer~\cite{wang2025unigrasptransformer}
% & 3200 & 1$^\dagger$ & 1 & 1 & $/$ &  &  & \checkmark &  & \checkmark \\
DexArt~\cite{bao2023dexart}
& 82 & 2 & 1 & $/$ &  &  & \checkmark & \checkmark &  \\
RLBench~\cite{james2019rlbench}
& * & 30 & 1 & $/$ &  & \checkmark &  & \checkmark &  \\
Robosuite~\cite{robosuite2020}
& 10 & 10 & 1 & $/$ &  & \checkmark &  & \checkmark &  \\
SIMPLER~\cite{li24simpler}
& 17 & 4 & - & $/$ & & \checkmark &  & \checkmark & \checkmark \\
RoboTwin 2.0~\cite{chen2025robotwin20}
& 731 & 11 & - & 100K &  & \checkmark &  & \checkmark & \checkmark \\
VLABench~\cite{zhang2024vlabench}
& 2164 & 11 & - & 5K &  & \checkmark &  & \checkmark & \checkmark \\
LIBERO~\cite{liu2023libero}
& 75 & * & 20 & 6.5K &  & \checkmark &  & \checkmark & \checkmark  \\
CALVIN~\cite{mees2022calvin}
& 30 & 10 & 4 & 20K &  &  \checkmark &  & \checkmark & \checkmark\\
BEHAVIOR-1K~\cite{BEHAVIOR}
& 9318 & 10 & 50 & 10K & \checkmark & \checkmark &  & \checkmark &  \\
Maniskill2~\cite{gu2023maniskill2}
& 2144 & 12 & * & 30K & \checkmark & \checkmark &  & \checkmark &  \\
RoboCasa~\cite{nasiriany2024robocasa}
& 2509 & 8 & 120 & 100K & \checkmark & \checkmark &  & \checkmark &  \\
UniDoorManip~\cite{li2024unidoormanip}
& 328 & 1 & 1 & $/$ & \checkmark & \checkmark &  & \checkmark & \checkmark \\
OWMM-Agent~\cite{chen2025owmmagent}
& 157 & 2 & 143 & 21K & \checkmark & \checkmark &  &  & \checkmark \\
GRUtopia~\cite{wang2024grutopia}
& 2956 & 2 & 100 & 900 & \checkmark & \checkmark &\checkmark &  &  \\
\textbf{MobileManiBench (Ours)} & \textbf{630} & \textbf{5} & \textbf{100} & \textbf{300K} & \checkmark & \checkmark & \checkmark & \checkmark & \checkmark \\

\bottomrule
\end{tabular}}
\vspace{2mm}
\caption{Comparison of existing robotic manipulation benchmarks and datasets. The top rows correspond to real-world datasets, while the bottom rows refer to simulation-based works. \textit{Skill Number}$^{\dagger}$ denotes the set of manipulation verbs included. \textit{Universal Model}$^{\dagger\dagger}$ indicates whether a universal policy has been trained within the benchmark. A ``–'' in \textit{Scene Number} indicates that the work uses a single tabletop layout with varying textures. A ``$/$'' in \textit{Trajectory Number} indicates that users must generate the trajectories themselves. An asterisk ``$*$'' denotes statistics not reported in the original paper. AgiBotWorld-Beta collects real-world G1-robot data through teleoperation, with only a small number of mobile-manipulation examples.}
\label{tab:benchmarks}
\vspace{-12mm}
\end{table*}

\noindent\textbf{Robotic Manipulation Benchmarks.}
Robotic manipulation~\cite{khazatsky2025droid,agibotworldcontributors2025agibotworld,zhao2025humanoideveryday,fu2024mobilealoha,wang2024trtm,bao2023dexart,james2019rlbench,chen2025robotwin20,zhang2024vlabench,gu2023maniskill2,nasiriany2024robocasa,li2024unidoormanip,chen2025owmmagent} remains a central challenge in embodied AI. Existing datasets and benchmarks can be broadly categorized into two groups, as summarized in Table~\ref{tab:benchmarks}. 
The first group comprises real-world datasets and evaluation protocols, such as DROID~\cite{khazatsky2025droid}, AgiBotWorld-Beta~\cite{agibotworldcontributors2025agibotworld} and Open X-Embodiment~\cite{o2024open}, which collect large-scale manipulation trajectories through extensive teleoperation or by aggregating data from multiple robot platforms. Despite their impressive scale, these works primarily focus on fixed-base or gripper-equipped robots performing mostly tabletop tasks. Recent efforts like DexGraspVLA~\cite{zhong2025dexgraspvla}, Humanoid Everyday~\cite{zhao2025humanoideveryday}, and Mobile ALOHA~\cite{fu2024mobilealoha} extend to dexterous hands and mobile manipulation, yet still rely on real-world human teleoperation, which limits their diversity and scalability. 
Overall, real-world data collection and model evaluation remain costly and risky, as they require complex scene setups and pose a high risk of hardware damage. Moreover, although existing real-world benchmarks cover a variety of manipulation skills, their evaluations are often limited to a narrow subset of tasks, leading to imbalanced training and testing coverage across object categories and skills.

The second group~\cite{bao2023dexart,james2019rlbench,robosuite2020,li24simpler,chen2025robotwin20,zhang2024vlabench,mees2022calvin,liu2023libero,gu2023maniskill2,nasiriany2024robocasa,li2024unidoormanip,chen2025owmmagent,wang2024grutopia,wang2025unigrasptransformer,RoboCerebra} alleviates the above limitation by leveraging high-fidelity simulation environments to simulate diverse manipulation tasks, collect large-scale trajectories, and perform extensive model evaluations. These methods typically collect data either through simulation-based teleoperation or by designing task-specific agents using rule-based methods or reinforcement learning (RL) policies.
While simulation greatly reduces the cost and risk, there remains a lack of large-scale, multi-modality trajectory datasets that support training universal models capable of generalizing across diverse object categories, manipulation skills, and realistic scenes. MobileManiBench follows this simulation-based paradigm, leveraging the high-fidelity \textit{NVIDIA Isaac Sim}~\cite{NVIDIA_Isaac_Sim} and a diverse set of digital assets to train universal RL policies across various robot–object–skill combinations. It  supports large-scale trajectory generation in diverse realistic digital scenes, facilitating scalable model training and evaluation for mobile-based robotic manipulation with both parallel grippers and dexterous hands.

\noindent\textbf{Vision-language-action Models.} VLA models~\cite{team2024octo,o2024open,li2024cogact,qu2025spatialvla,black2410pi0,intelligence2025pi_,kim2025fine,brohan2022rt,li2022vision,zitkovich2023rt,liu2024rdt,wen2025dexvla,bjorck2025gr00t,wang2025unified, zhong2025dexgraspvla, cheang2025gr, li2025scalable,team2025gemini,yang2025magma,shi2025memoryvla,zhang2025dreamvla} have emerged as a promising framework for learning universal robotic policies that generalize across diverse objects, tasks, and scenes. Recent works~\cite{black2410pi0,intelligence2025pi_,li2024cogact,wen2025dexvla,bjorck2025gr00t} advance this direction by integrating vision–language (VL) backbones~\cite{beyer2024paligemma,chen2023pali,driess2023palm,liu2023visual,team2023gemini,karamcheti2024prismatic} with diffusion-based or flow-matching-based action modules. By combining the pretrained VL priors, multi-modality representations, and continuous action modeling, these approaches substantially lead the robotic manipulation performance.

However, existing VLA models are mostly trained and evaluated on fixed-base, gripper-equipped robots performing tabletop manipulation tasks, with limited examples of dexterous hands or mobile-based manipulation~\cite{team2024octo,o2024open,li2024cogact,qu2025spatialvla,kim2025fine,brohan2022rt,li2022vision,zitkovich2023rt,liu2024rdt,wen2025dexvla,wang2025unified, zhong2025dexgraspvla, cheang2025gr,team2025gemini,yang2025magma,shi2025memoryvla,zhang2025dreamvla}. 
MobileManiBench extends this research by providing large-scale synthetic trajectories for mobile-based robots equipped with either parallel grippers or dexterous hands, covering diverse objects, tasks, and scenes. It systematically studies the effect of different model structures and input modalities on mobile manipulation performance, filling a critical gap in current VLA research.

\vspace{-2mm}
\section{Problem Formulation}
\label{sec:problem}
\vspace{-1mm}

\noindent\textbf{Mobile-based Robot.} 
MobileManiBench employs two mobile-based robots: the G1 robot from AGIBOT~\cite{agibot} and the XHand robot from RobotEra~\cite{xhand}, as shown in Figure~\ref{fig:teaser}. Although both robots are equipped with dual arms, we fix the left arm and activate only the right arm for manipulation. Each robot has a 7-DOF right arm and a 2-DOF mobile base (one rotational DOF around the z-axis and one translational DOF along the y-axis). The G1 robot uses a parallel gripper with 1 DOF as its end effector, whereas the XHand robot is equipped with a dexterous 12-DOF hand. Both robots are controlled using (6+D) dimensional actions, where the first 6 dimensions represent the wrist pose displacement relative to the previous frame, from which inverse kinematics computes the target joint angles of mobile base and right arm. The remaining D dimensions correspond to the target joint angles of end effector. This formulation yields 7-d actions for the G1 robot and 18-d actions for the XHand robot.

\noindent\textbf{Mobile-based Manipulation Objects, Skills, and Tasks.}
MobileManiBench encompasses both the articulated and holistic objects sourced from the PartNet-Mobility~\cite{xiang2020partnetmobility}, UniDoorManip~\cite{li2024unidoormanip}, and YCB~\cite{calli2015ycb} datasets, totaling 630 objects across 20 categories (holistic objects such as apples, bottles, and toys are grouped into a single ``holistic'' category). We define five mobile manipulation skills: \textit{open} and \textit{close} for objects including boxes, toilets, laptops, trashcans, dishwashers, ovens, microwaves, refrigerators, washing machines, cars, safes, fridges, cabinets, windows, lever doors, round doors, faucets, and tables; \textit{pull} and \textit{push} for carts; and \textit{pick} for holistic (YCB) objects. This formulation yields over 100 distinct tasks for large-scale data collection and model training, including tasks such as opening a laptop, closing a cabinet, pushing a cart, pulling a cart, picking an apple, and picking a mustard bottle.

\vspace{-3mm}
\section{Method}
\label{sec:method}
\vspace{-3mm}

\textbf{Overview.} The construction of MobileManiBench consists of two stages. First, in the \textit{MobileManiRL Training} stage (Section~\ref{sec:MobileManiRL}), we train a state-based reinforcement learning (RL) policy for each robot–object–skill combination, covering robots equipped with either a parallel gripper or a dexterous hand. Second, in the \textit{MobileManiDataset Generation} stage (Section~\ref{sec:MobileManiDataset}), each trained RL policy is deployed in diverse digital scenes to collect successful manipulation trajectories. Each trajectory includes a natural language instruction paired with synchronized data: multi-view camera images, object states, robot states, and actions. 

Building on \textit{MobileManiDataset}, we further introduce \textit{MobileManiVLA} (Section~\ref{sec:MobileManiVLA}), where all trajectories across robot–object–skill–scene combinations are aggregated to train a universal vision-language-action model for each robot, enabling strong generalization to unseen objects and unseen scenes.
\vspace{-4mm}

\begin{figure}[h]
    \centering
    \includegraphics[width=0.6\linewidth]{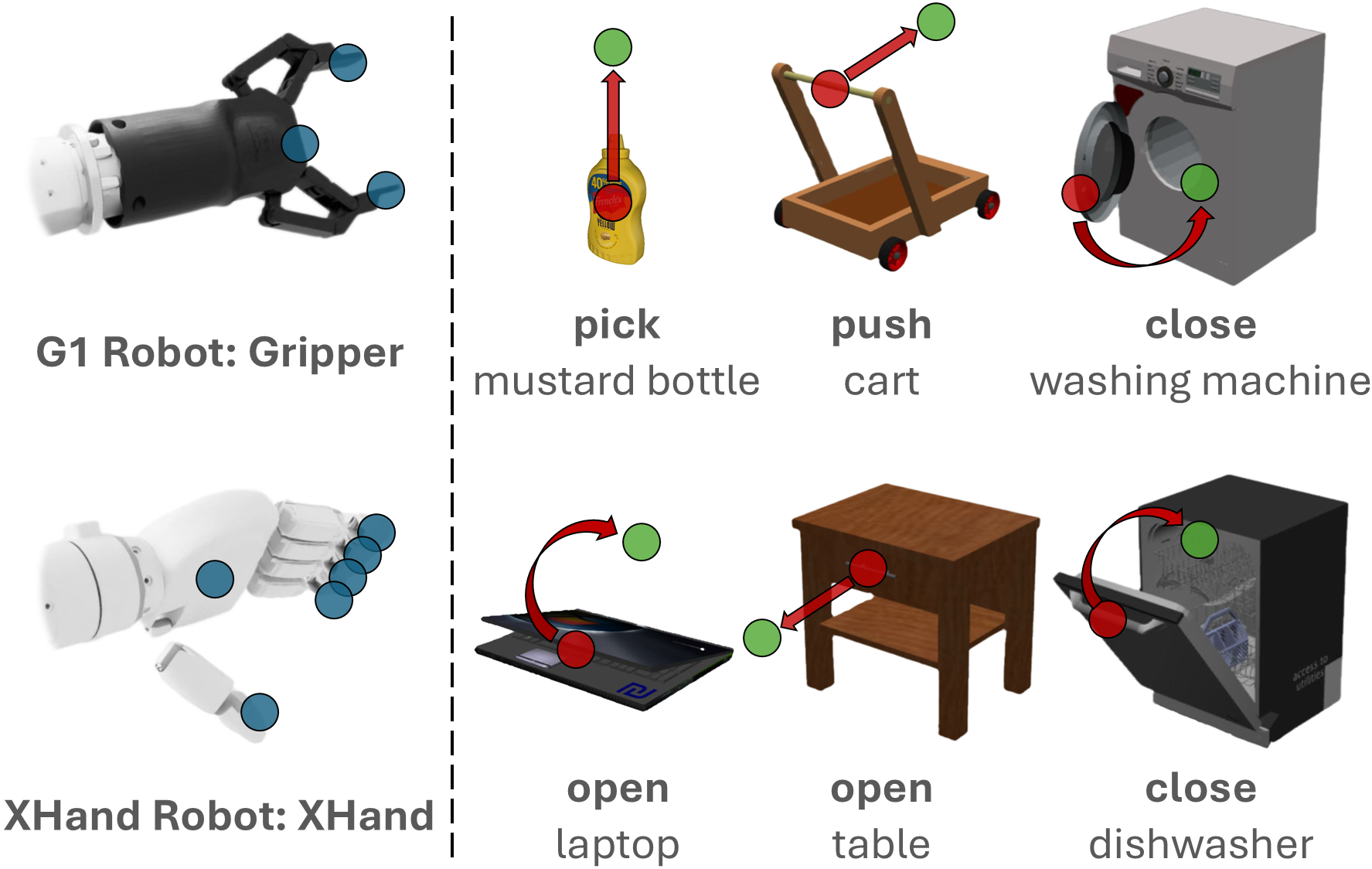}
    \vspace{-2mm}
    \caption{
    Definitions of the robot gripper/hand points (blue), object grasp point (red), and goal point (green) across diverse tasks.
    }
    \vspace{-12mm}
    \label{fig:points}
\end{figure}

\vspace{-0.5mm}
\subsection{MobileManiRL Training}
\label{sec:MobileManiRL}
\vspace{-0.5mm}
Given that our benchmark involves 2 mobile-based robots, 630 objects across 20 categories, 5 manipulation skills, and over 100 tasks, teleoperating or manually designing policies for each configuration would be prohibitively time-consuming. To address this, we propose a universal state-based reinforcement learning (RL) policy, termed \textit{MobileManiRL}, which parameterizes each robot–object–skill combination using keypoint-based displacements of the robot gripper/hand points, the object grasp point, and the goal point~\cite{huang2024rekep,zhang2024graspxl}, as illustrated in Figure~\ref{fig:points}. A universal reward function $R$ encourages the robot’s gripper/hand points to reach the object grasp point and transport it to the goal point. This formulation enables a single RL policy to generalize across diverse manipulation scenarios while maintaining task-specific success.

\begin{table}[h]
    \centering
    %\small
    \begin{tabular}{l|cc}
        \toprule
Input Type  & G1 Robot & XHand Robot \\
 % Method & (Online) & (Offline) \\
\midrule
Time & 30 & 30 \\
Object State & 9 & 9 \\
Robot Proprioception & 78 & 135 \\
Robot-Object Distance & 22 & 31 \\
Previous Action & 7 & 18 \\
    \bottomrule
    \end{tabular}
    \vspace{2mm}
    \caption{Dimensions of the input types used in MobileManiRL for the G1 and XHand robots. Detailed definitions of each input element are provided in the Appendix.}
    \label{tab:rl_input}
    \vspace{-10mm}
\end{table}

\noindent\textbf{Inputs.}
Table~\ref{tab:rl_input} summarizes the five input types used in MobileManiRL. Specifically, \textit{Time} encodes the timestep embedding; \textit{Object State} represents the states of the object grasp point and goal point; \textit{Robot Proprioception} captures the robot’s body and joint states; \textit{Robot–Object Distance} measures the distances between the robot gripper/hand points and the object grasp point; and \textit{Previous Action} records the previous (6+D)-dimensional action.

\noindent\textbf{Network Architecture.} Each MobileManiRL policy network is a 4-layer MLP with hidden dimensions of \{1024, 1024, 512, 512\}, followed by an action prediction head implemented as a single fully connected layer. This head outputs a (6+D) dimensional vector (7-d for the G1 robot and 18-d for the XHand robot) representing the action at the current time step. The value network adopts the same architecture as the policy network but outputs a single scalar.

\noindent\textbf{Reward Function.} 
The reward function $R$ is designed as:
\begin{equation}
\vspace{-1mm}
\label{eq:reward}
R = R_{d} + (1-f_{g})R_{a} + f_{g}(R_{g} + R_{m} + R_{s}),
\vspace{-1mm}
\end{equation}
where the global distance reward $R_{d}$ penalizes the distances between the gripper/hand points and the object grasp point, as well as between the object grasp point and the goal point, encouraging the robot to grasp the object and move it toward the goal. The grasp flag $f_{g}$ is set to 1 when the distance between the gripper/hand points and the object grasp point falls below a predefined threshold. The approach action reward $R_{a}$ guides the gripper/hand toward the object grasp point via a reference approach action until a grasp is achieved. Once grasped (i.e., $f_{g}=1$), the grasp reward $R_{g}$ provides a bonus; the move action reward $R_{m}$ encourages the gripper/hand to move to the goal point via a reference move action; finally the success reward $R_{s}$ provides an extra bonus when the object reaches the goal. Formal definitions of all reward elements are provided in the Appendix. A manipulation is considered successful if the object grasp point reaches the goal point within $T = 300$ steps.

\begin{figure}[!t]
    \centering
    \includegraphics[width=0.65\linewidth]{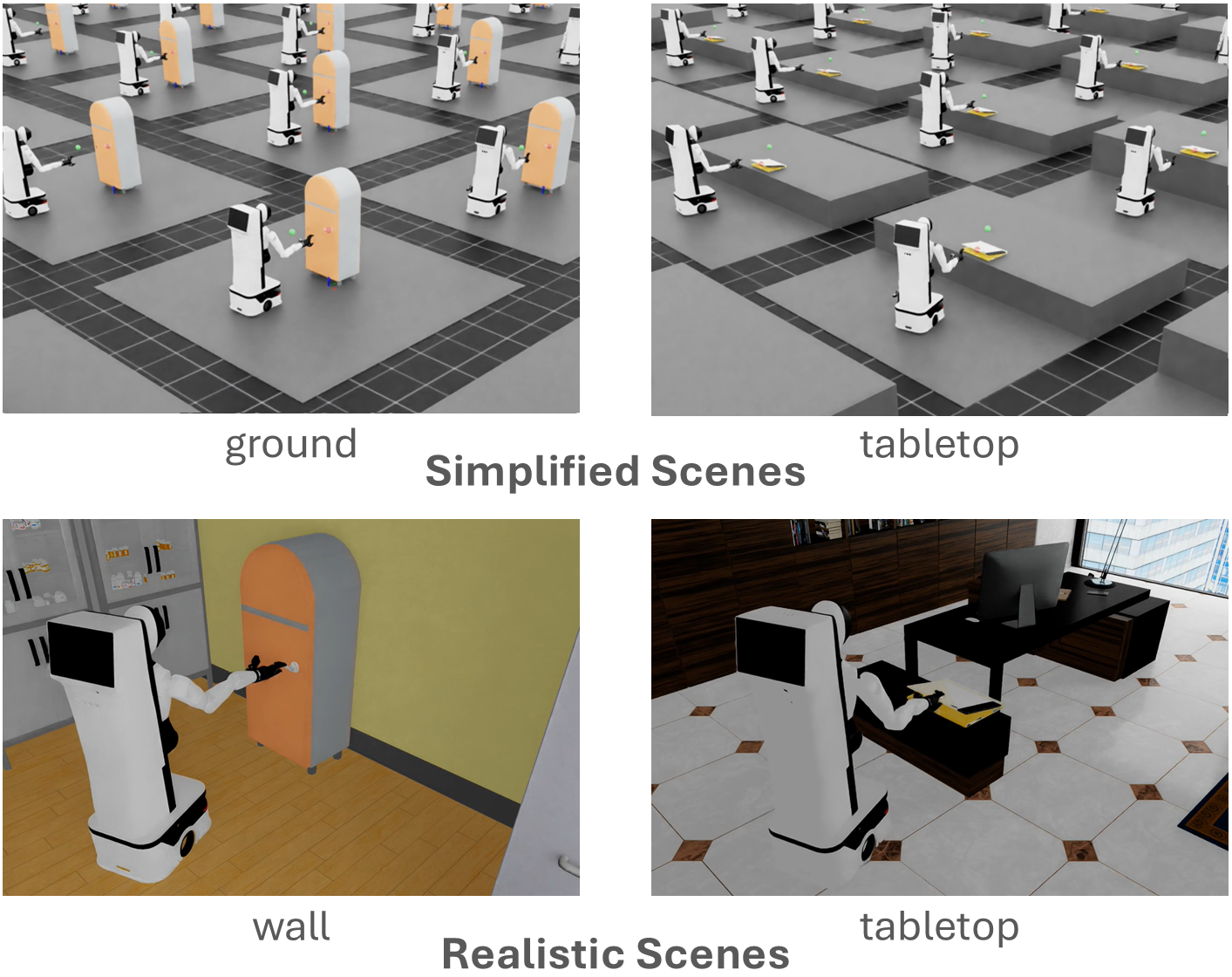}
    \vspace{-2mm}
    \caption{
    Illustrations of simplified scenes for MobileManiRL training and realistic scenes for MobileManiDataset generation and MobileManiVLA evaluation.
    }
    \vspace{-6mm}
    \label{fig:scenes}
\end{figure}

\noindent\textbf{Simplified Scene Training.} 
We place the 20 categories of objects into 2 simplified scene settings in Isaac Sim and train one MobileManiRL policy for each robot-object-skill combination, as shown in Figure~\ref{fig:scenes}. The \textit{ground} scene includes toilet, trashcan, refrigerator, washing machine, car, fridge, cabinet, window, lever door, round door, table, and cart; while the \textit{tabletop} scene includes box, laptop, dishwasher, oven, microwave, safe, faucet, and holistic (YCB) objects. Each object is initialized either on the ground or on the tabletop with random heights, facing the negative y axis, and with its grasp point's x-y coordinates set to zero. The robot mobile base is initialized facing the object grasp point with random positional offsets ($1.0m$ to $1.5m$ along the x and y axes) and rotational offsets (${-15}^{o}$ to ${15}^{o}$ around the z axis) to enhance policy robustness. Under this setup, we train MobileManiRL on each of the 1,182 robot-object–skill combinations for both G1 robot and XHand robot, achieving mean success rates of \textbf{89.6\%} and \textbf{92.9\%}, respectively.

\vspace{-1mm}
\subsection{MobileManiDataset Generation}
\label{sec:MobileManiDataset}
\vspace{-1mm}

\noindent\textbf{Realistic Scene Rendering.} 
Each trained MobileManiRL policy is now able to perform its assigned robot-object-skill across diverse robot initial poses. We then place the 20 categories of objects into 5 realistic scene settings for trajectory rendering: space (cart), wall (toilet, trashcan, refrigerator, washing machine, fridge, cabinet, table), door (window, lever door, round door), outdoor (car), and tabletop (box, laptop, dishwasher, oven, microwave, safe, faucet, holistic objects). For each scene setting, we manually annotate 20 scene placements using digital assets from the Isaac Sim~\cite{NVIDIA_Isaac_Sim} and Genie Sim~\cite{contributors2025geniesimrepo}, yielding a total of 80 seen scene placements for VLA training and 20 unseen reserved for testing. These scenes cover a variety of everyday environments, including kitchen, bedroom, hospital, office, warehouse, and parking lot, as shown in the Appendix.

\noindent\textbf{Dataset Analysis.}
For each of the G1 robot and the XHand robot, MobileManiDataset comprises 630 objects across 20 categories, manipulated through 5 skills and over 100 tasks using 1,182 robot-object-skill combinations of MobileManiRL, which are further distributed across 100 scene placements. The dataset is split into 506 objects and 80 scenes for VLA training, plus 124 objects and 20 scenes for testing, yielding a total of 15,232 robot-object–skill–scene combinations for training and 920 combinations for testing. For each training combination, we generate 10 successful manipulation trajectories in Isaac Sim, resulting in 150K training trajectories. Each manipulation trajectory, $\mathcal{T}=\{L, (I_1, S_1,A_1),\dots, (I_t,S_t,A_t), \dots, (I_T,S_T,A_T)\}$, is recorded at 30 FPS with an average length of 160 frames, including one natural language instruction $L$; synchronized $520\times520$ RGB, depth, and segmentation images $I_t$ from both head-view and wrist-view cameras; the corresponding object and robot states $S_t$; and the executed (6+D) dimensional action $A_t$ at each timestep $t$. All states and actions are recorded in the global world coordinate frame.

\begin{figure}[!t]
    \centering
    \includegraphics[width=0.9\linewidth]{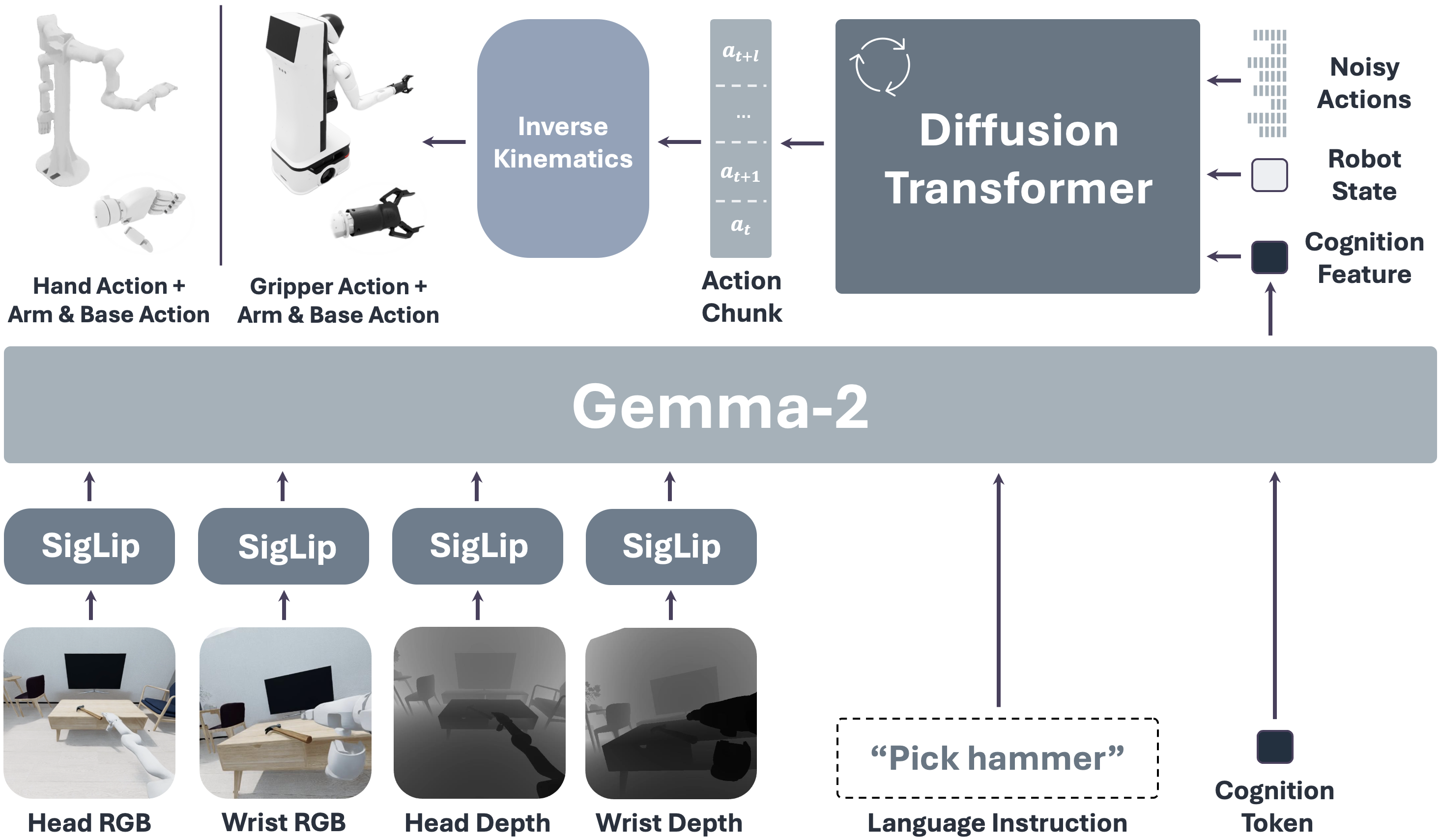}
    \vspace{-2mm}
    \caption{Model architecture of MobileManiVLA with multi-modality inputs.}
    \vspace{-4mm}
    \label{fig:vla}
\end{figure}

\subsection{MobileManiVLA Training}
\label{sec:MobileManiVLA}

The objective is to use the generated MobileManiDataset to train a universal vision-language-action model, MobileManiVLA, for each of the G1 robot and XHand robot. MobileManiVLA is capable of performing mobile-based robotic manipulation of diverse ground and tabletop objects across varying initial poses, and can generalize to both unseen objects and unseen scenes. Following the design of CogACT~\cite{li2024cogact}, we structure our MobileManiVLA into three components: the vision, language, and action modules, as shown in Figure~\ref{fig:vla}.

\noindent\textbf{Vision and Language Modules.}
The vision and language modules are initialized from a pretrained vision–language model, PaliGemma-2~\cite{steiner2024paligemma}. The vision module processes multi-view image inputs, including two RGB images and two depth images captured from the head-view and wrist-view cameras (each reshaped to \(224 \times 224 \times 3\)). These images are encoded via a SigLIP vision encoder into dense visual embeddings. In parallel, a language instruction of the form “\textless skill\textgreater\ \textless object\textgreater” (such as “open faucet” or “close door”) is tokenized and projected into language embeddings. The vision and language embeddings are then fused via a linear projection for alignment. Together with a learnable \textit{cognition token} $c$, they are fed into a Gemma-2 language model~\cite{team2024gemma} to produce an output feature $f_t^c$ that captures both perceptual and instructional context.

\noindent\textbf{Action Module with State Conditions.}
The action module is implemented as a diffusion-transformer (DiT) that takes as input: 1) a series of noisy actions $(a_t^i, a_{t+1}^i, \dots, a_{t+N}^i)$, where $i$ denotes the current denoising step; 2) the encoded cognition feature $f_t^c$; and 3) the state feature $f_t^s$, obtained by encoding the 6-d robot wrist pose using a lightweight MLP.
Conditioned on $f_t^c$ and $f_t^s$, the DiT iteratively predicts the clean actions $(a_t, a_{t+1}, \dots, a_{t+N})$ over multiple denoising steps, progressively refining its predictions toward physically consistent and goal-directed motions.
All states and actions are expressed in the robot mobile base coordinate frame for consistency across scenes.

\noindent\textbf{Training Objective and Inference Strategy.}
The vision, language, and action modules are trained end-to-end by minimizing the mean squared error (MSE) between the predicted and the ground truth Gaussian noises, with the loss function defined as:

\begin{equation}
\mathcal{L}_{\text{MSE}} = \mathbb{E}_{\epsilon \sim \mathcal{N}(0,1),\, i}\|\hat{\epsilon}^i - \epsilon\|_2,
\end{equation}
where $\hat{\epsilon}^i$ is the predicted noise for the noisy action sequence $(a_t^i, a_{t+1}^i, \dots, a_{t+N}^i)$ at the $i$-th denoising step, and $\epsilon$ is the corresponding ground truth Gaussian noise.

During inference, MobileManiVLA predicts a chunk of actions with length $N=16$ conditioned on the current instruction and observations. we adopt the adaptive ensemble strategy proposed in CogACT~\cite{li2024cogact} to improve trajectory smoothness and robustness, with a window size of $K=4$.

\section{Experiment}
\label{sec:experiment}

For each of the G1 robot and XHand robot, we first train MobileManiRL on each of the 1,182 robot-object-skill combinations, with 1,024 simulation environments, a learning rate of $1e^{-3}$, and 4K iterations. Training is conducted in parallel on 32 NVIDIA V100 GPUs and requires about 4 days per robot.
We then collect 15,232 robot-object-skill-scene combinations for training and 920 combinations for testing to generate the MobileManiDataset, resulting in 150K training trajectories. Trajectory generation is conducted in parallel on 8 NVIDIA RTX A6000 GPUs and requires about 6 days per robot.
Finally, We train the MobileManiVLA model on MobileManiDataset, with a batch size of 480, a learning rate of $2e^{-5}$, and 320K iterations. Training is conducted on 8 NVIDIA B200 GPUs and takes about 12 days per robot.

\begin{figure*}[h]
    \centering
    \vspace{-4mm}
    \includegraphics[width=1.0\linewidth]{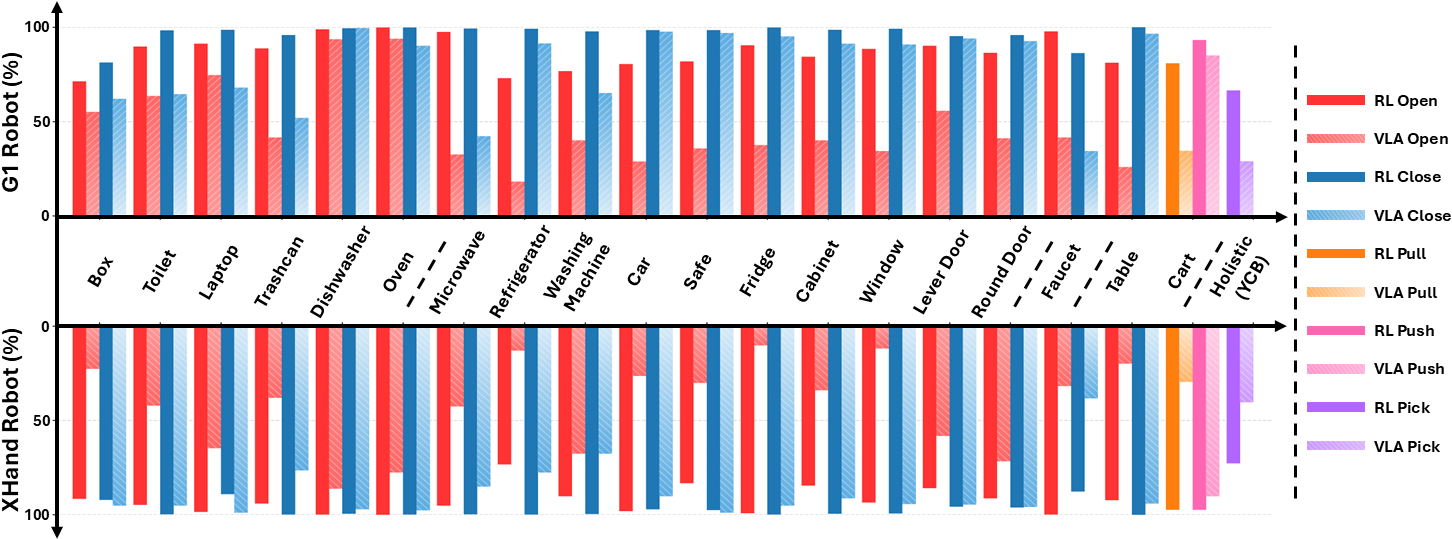}
    \vspace{-2mm}
    \caption{Success rates of MobileManiRL and MobileManiVLA on the G1 robot and XHand robot across 20 object categories and 5 mobile manipulation skills. In terms of manipulation motion patterns, objects like box and oven require lid-flipping upward or downward; car, fridge, and door require door-handle grasping followed by pivoting left or right; faucet requires handle rotation; table and cart require handle grasping with pulling or pushing; holistic (ycb) objects require object grasping and lifting.}
    \vspace{-10mm}
    \label{fig:object_success}
\end{figure*}

\subsection{Main Results}
\vspace{-2mm}
Figure~\ref{fig:object_success} demonstrates the success rates of MobileManiRL and MobileManiVLA across the 20 object categories, while Table~\ref{tab:skill_success} summarizes the mean success rates over the 5 mobile manipulation skills. MobileManiRL is evaluated on seen objects with 1024 episodes per robot-object-skill combination, whereas MobileManiVLA is evaluated on unseen objects and scenes, with 10 episodes per robot-object-skill-scene combination. Both experiments are conducted with randomized robot initial poses to assess robustness.

\begin{table}[h]
%\small
%\setlength\tabcolsep{5pt}
    \centering
    \vspace{-4mm}
    % \resizebox{\linewidth}{!}{
    \begin{tabular}{l|cc|cc}
        \toprule
        \multirow{2}{*}{Skill} & \multicolumn{2}{c|}{MobileManiRL (\%)} & \multicolumn{2}{c}{MobileManiVLA (\%)} \\  
        \cmidrule(lr){2-3} \cmidrule(lr){4-5} 
        & \: G1 Robot \: & XHand Robot \: & \: G1 Robot \: & XHand Robot \: \\
        \midrule
        Open & 86.6 & \textbf{91.9} & \textbf{42.9} & 34.5 \\
        Close & 96.2 & \textbf{96.5} & 75.8 & \textbf{77.5} \\
        Pull & 80.8 & \textbf{97.3} & \textbf{34.4} & 29.4 \\
        Push & 93.1 & \textbf{97.2} & 85.1 & \textbf{90.0} \\
        Pick & 66.4 & \textbf{72.6} & 28.8 & \textbf{40.2} \\
        \midrule
        Mean & 89.6  & \textbf{92.9}  & 56.7  &  \textbf{57.3} \\
        \bottomrule
    \end{tabular}
    \vspace{2mm}
    \caption{Success rates of MobileManiRL and MobileManiVLA on the G1 robot and XHand robot across 5 manipulation skills.}
    \label{tab:skill_success}
    \vspace{-10mm}
\end{table}

\noindent\textbf{Takeaway 1 — Object structure and skill complexity drive performance variance.}
For both MobileManiRL and MobileManiVLA, object categories like toilet, laptop, and dishwasher achieve higher success rates due to their relatively simple structures and motion patterns, typically involving lid-flipping without requiring complex handle grasping.
In contrast, object categories such as car, door, and table present greater challenges, as they demand precise handle localization and stable grasp control.
Regarding the mobile manipulation skills, open, pull, and pick, which require accurate and stable grasping of the object, are generally more difficult to learn than close and push.

\noindent\textbf{Takeaway 2 — Generalization remains challenging for VLA models with the implicit inputs.}
Compared with MobileManiRL, the success rates of MobileManiVLA exhibit a moderate decline for two primary reasons. First, MobileManiRL is trained and evaluated individually on each seen robot–object–skill combination, whereas MobileManiVLA is trained as a universal model and evaluated on unseen objects and unseen scenes, which increases diversity and generalization difficulty. Second, MobileManiRL leverages explicit state-based observations, such as the world positions of the robot gripper/hand points, the object grasp and goal points, providing precise geometric information. In contrast, MobileManiVLA relies solely on implicit sensory inputs, including a language instruction, multi-view RGB-D images, and the robot wrist pose represented in the mobile base coordinate frame. This transition from explicit to implicit observations introduces additional learning challenges but enables MobileManiVLA to generalize effectively to unseen objects and unseen scenes in both simulation and the real world. The real world inference of MobileManiVLA is demonstrated in the Appendix.

\noindent\textbf{Takeaway 3 — Dexterous hands improve manipulation precision and versatility.}
As shown in Table~\ref{tab:skill_success}, with the MobileManiRL, the XHand robot achieves a higher success rate of \textbf{92.9\%}, compared to \textbf{89.6\%} for the G1 robot, highlighting the advantages of dexterous manipulation, particularly on the open, pull, and pick skills. 
While with MobileManiVLA, the G1 robot and XHand robot achieve comparable success rates of \textbf{56.7\%} and \textbf{57.3\%}, respectively. In this setting, the XHand robot performs worse on the open and pull skills, where its dexterous fingers often collide with surrounding object surfaces, preventing stable handle grasps. Such disturbance between the object body (e.g., door faces) and the dexterous fingers significantly degrades the VLA performance. Nevertheless, with MobileManiVLA , the XHand robot outperforms the G1 robot on the pick skill, where the dexterous fingers can easily establish the force-closure grasps on the holistic objects.

\subsection{Ablation Studies and Model Comparisons}
We conduct ablation studies and model comparisons on a subset of MobileManiDataset that includes only the challenging skills: open, pull, and pick for the G1 robot, covering 272 training and 66 testing objects across 6 categories: laptop, cabinet, faucet, table, cart and holistic objects, providing 4,352 robot-object-skill-scene combinations for training and 264 unseen combinations for testing.
All VLA models are trained with a batch size of 120 and 160K iterations, which are evaluated on unseen objects and scenes with 10 episodes per combination.

\begin{table*}[h]
%\small
%\setlength\tabcolsep{5pt}
    \centering
    % \resizebox{\linewidth}{!}{
    \resizebox{0.99\textwidth}{!}{\begin{tabular}{cccc|ccc|c}
        \toprule
        \multicolumn{4}{c|}{Image Inputs} & \multicolumn{3}{c|}{State Inputs} & \multirow{2}{*}{Success Rate} \\
        \cmidrule(lr){1-4} \cmidrule(lr){5-7} 
        Head RGB & Head Depth & Wrist RGB & Wrist Depth & Wrist Pose & Grasp Point & Goal Point & (\%) \\
        \midrule
        \checkmark &  &  &  & \checkmark &  &  & 7.9 \\
        \checkmark & \checkmark &  &  & \checkmark &  &  & 14.1 \\
        \checkmark &  & \checkmark &  & \checkmark &  &  & 14.9 \\
        \checkmark & \checkmark & \checkmark & \checkmark & \checkmark &  &  & \textbf{28.2} \\
        \cmidrule(lr){1-4} \cmidrule(lr){5-8} 
        % \checkmark &  &  &  &  &  &  & 6.5 \\
        \checkmark & \checkmark & \checkmark & \checkmark &  &  &  & 22.4 \\
        \checkmark & \checkmark & \checkmark & \checkmark & \checkmark & \checkmark &  & 32.3 \\
        \checkmark & \checkmark & \checkmark & \checkmark & \checkmark & \checkmark & \checkmark & \textbf{36.6} \\  
        \bottomrule
    \end{tabular}}
    \vspace{2mm}
    \caption{Ablation studies of MobileManiVLA on the G1 robot with different image inputs (top rows) and state inputs (bottom rows).}
    \label{tab:ablation_inputs}
    \vspace{-6mm}
\end{table*}

\noindent\textbf{Takeaway 4 — Multi-view and multi-modality visual inputs significantly enhance policy generalization.}
Table~\ref{tab:ablation_inputs} shows the success rates of MobileManiVLA with different image inputs. 
Using only the head-view RGB image yields a success rate of only 7.9\%, indicating that a single global view is insufficient for precise manipulation in mobile-based settings.
Adding either the head-view depth image or the wrist-view RGB image improves the performance to 14.1\% and 14.9\% respectively, suggesting that additional geometric or visual cues enhance spatial understanding.
Finally, combining both RGB-D images from the head-view and wrist-view cameras achieves the highest success rate of 28.2\%, confirming that multi-view and multi-modality visual inputs significantly enhance manipulation accuracy and generalization ability.

\noindent\textbf{Takeaway 5 — State inputs from robot proprioception or simple object detection further improve performance.}
Table~\ref{tab:ablation_inputs} reports the impact of different state inputs on the performance of MobileManiVLA. Compared with no state input, encoding the robot wrist pose in its mobile base coordinate frame increases the success rate from 22.4\% to 28.2\%, highlighting the importance of proprioceptive information for precise manipulation. Further incorporating the pseudo-observations, such as the first-frame positions of the object grasp point and goal point expressed in the robot mobile base coordinate frame, leads to additional performance gains, confirming that even coarse spatial priors about the object can enhance policy grounding and performance.

\begin{table}[h]
    \centering
    \begin{tabular}{cccc|c}
        \toprule
        Uneen & Unseen & Seen & Seen & \multirow{2}{*}{Success Rate (\%)} \\ 
        Object & Scene & Object & Scene & \\ \midrule
         &  & \checkmark & \checkmark & 59.6 \\
         & \checkmark & \checkmark &  & 51.3 \\
        \checkmark &  &  & \checkmark & 39.2 \\
        \checkmark & \checkmark &  &  & 28.2 \\
        \bottomrule
    \end{tabular}
    \vspace{2mm}
    \caption{Ablation studies of MobileManiVLA on the G1 robot with seen or unseen objects and scenes.}
    \label{tab:ablation_unseen_items}
    \vspace{-10mm}
\end{table}

\begin{table}[h]
    \centering
    \begin{tabular}{l|c}
        \toprule
        Model & Success Rate (\%)  \\
        \midrule
        OpenVLA~\cite{kim2024openvla} (7B) & 4.5 \\
        CogACT~\cite{li2024cogact} (7B) & 6.8 \\
        $\pi_0$~\cite{black2410pi0} (3B) & 11.2 \\
        $\pi_{0.5}$~\cite{intelligence2025pi_} (3B) & 18.8 \\
        MobileManiVLA (3B) & \textbf{28.2} \\
        \bottomrule
    \end{tabular}
    \vspace{2mm}
    \caption{Model comparisons of representative VLA models with their default architectures and input modalities on the G1 robot.}
    \label{tab:model_comparisons}
    \vspace{-10mm}
s\end{table}

\noindent\textbf{Takeaway 6 — Unseen objects present greater challenges than unseen scenes.}
Table~\ref{tab:ablation_unseen_items} indicates the success rates of MobileManiVLA on seen or unseen objects and scenes. The model achieves the highest success rate of 59.6\% when both objects and scenes are seen during testing. When evaluated with seen objects in unseen scenes or unseen objects in seen scenes, the success rates drop to 51.3\% and 39.2\% respectively, indicating that generalizing to unseen object structures presents greater challenges than adapting to new scene layouts. The lowest success rate of 28.2\% occurs when both objects and scenes are unseen, reflecting the compounded difficulty of transferring manipulation skills to entirely novel environments.

\noindent\textbf{Takeaway 7 — MobileManiBench provides a unified platform for training and evaluating VLA models.}
By offering standardized training and evaluation protocols across diverse robots, objects, tasks, scenes and input modalities, MobileManiBench enables consistent benchmarking of existing VLA models with their default architectures and inputs, as shown in Table~\ref{tab:model_comparisons}. OpenVLA and CogACT receive only head-view RGB image, resulting in low success rates of 4.5\% and 6.8\%, respectively. $\pi_0$ and $\pi_{0.5}$ process both head-view and wrist-view RGB images together with wrist pose states, achieving improved success rates of 11.2\% and 18.8\%, respectively. The stronger performance of $\pi_{0.5}$ arises from its pretraining on more mobile manipulation trajectories. Overall, MobileManiVLA achieves the highest success rate of 28.2\% by jointly leveraging head-view and wrist-view RGB-D images along with wrist pose states, highlighting the advantages of multi-view and multi-modality inputs.

\begin{table}[h]
    \centering
    \begin{tabular}{cc|c}
        \toprule
        Fixed Base & Mobile Base & Success Rate (\%)  \\
        \midrule
         & \checkmark & \textbf{82.8} \\
        \checkmark &  & 25.4 \\
        \bottomrule
    \end{tabular}
    \vspace{2mm}
    \caption{Success rates of MobileManiRL on the G1 robot with fixed or mobile base.}
    \label{tab:ablation_mobile_base}
    \vspace{-5mm}
\end{table}

\noindent\textbf{Takeaway 8 — Base mobility is crucial for effective manipulation beyond tabletop tasks.}
To evaluate the importance of base mobility during manipulation, we fix the G1 robot base and initialize it closer to the object grasp point than before ($0.5m$ to $1.0m$ along the x and y axes). We train MobileManiRL under this fixed-base setting on several robot-object-skill combinations, including open laptop, open cabinet, open faucet, open table, pull cart and pick holistic (YCB) objects. As shown in Table~\ref{tab:ablation_mobile_base}, the success rates of MobileManiRL drop drastically when the robot base is fixed, confirming that base mobility is essential for spatial manipulation beyond static tabletop interactions.

\section{Conclusion}
\label{sec:conclusion}
We introduce MobileManiBench, a large-scale simulation-based benchmark designed to simplify model verification for mobile-based robotic manipulation. MobileManiBench encompasses 2 mobile-based robots with either parallel grippers or dexterous hands, 630 objects spanning 20 categories, and 5 mobile manipulation skills across 100+ tasks. By training a universal MobileManiRL policy on each robot-object-skill triplet, the benchmark autonomously generates the MobileManiDataset in 100 realistic scenes, which contains 300K trajectories with 3 data modalities: language instructions, multi-view RGB-depth-segmentation images, synchronized object/robot states and actions. Experiments demonstrate that MobileManiBench serves as a unified and scalable platform for developing and evaluating next-generation vision–language–action models, accelerating research toward general-purpose, embodied intelligence.

\bibliographystyle{splncs04}
\bibliography{main}
\end{document}

\title{Supplementary Material for MobileManiBench \vspace{-5mm}} 

\titlerunning{Abbreviated paper title}

% % TODO FINAL: Replace with your author list. 
% % Include the authors' OCRID for the camera-ready version, if at all possible.
% Second Author\inst{2,3}\orcidlink{1111-2222-3333-4444} \and
% Third Author\inst{3}\orcidlink{2222--3333-4444-5555}}

% % TODO FINAL: Replace with an abbreviated list of authors.

% % TODO FINAL: Replace with your institution list.
% Springer Heidelberg, Tiergartenstr.~17, 69121 Heidelberg, Germany

\maketitle

\vspace{-4mm}
\section{Implementation Details}
\label{sec:method_supp}
\vspace{-8mm}
\begin{figure}[h]
    \centering
    \includegraphics[width=0.7\linewidth]{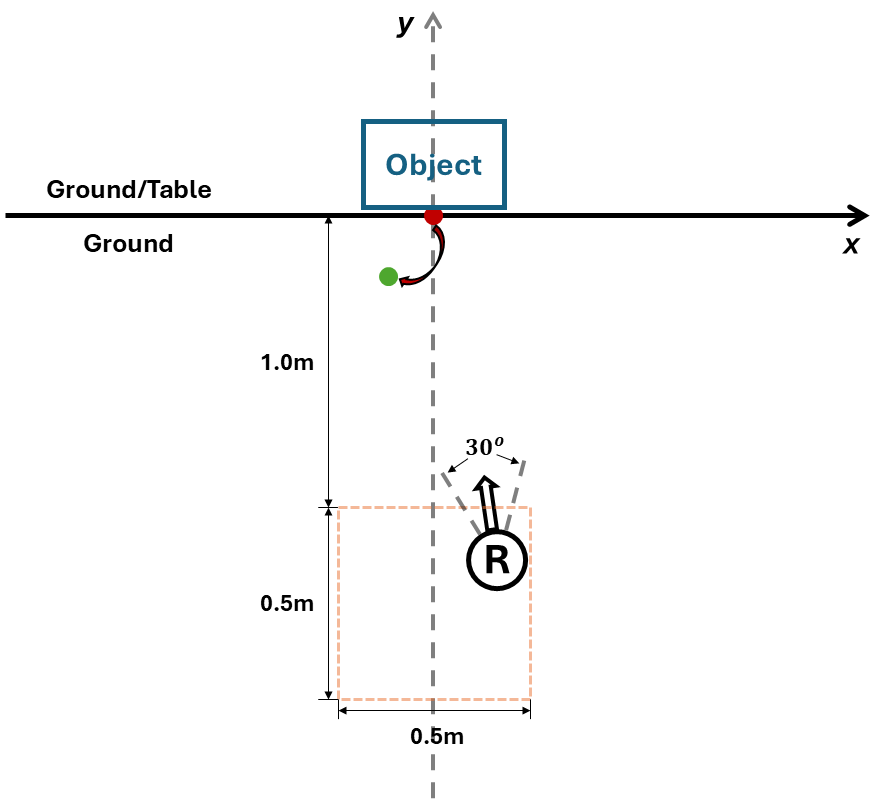}
    \vspace{-1mm}
    \caption{
        Initialization of the robot, object, ground, and table.
    }
    \vspace{-4mm}
    \label{fig:init}
\end{figure}

\vspace{-4mm}
\subsection{Digital Assets}

MobileManiBench utilizes NVIDIA Isaac Sim 4.5~\cite{NVIDIA_Isaac_Sim} as the simulation platform and employs NVIDIA Isaac Lab~\cite{NVIDIA_Isaac_Lab} to construct the simulation environments. All policies operate at 30 FPS, while the simulation runs at 60 FPS.

\noindent\textbf{Robot Assets.}
MobileManiBench encompasses 2 digital robots: the G1 robot from AGIBOT~\cite{agibot} and the XHand robot from RobotEra~\cite{xhand}. The USD files for these robots are created by processing their original URDF files and adding 2 degrees of freedom (DOFs) at their robot base: one rotational DOF around the z-axis and one translational DOF along the y-axis.

\noindent\textbf{Object Assets.}
MobileManiBench includes 630 digital objects across 20 categories: box, toilet, laptop, trashcan, dishwasher, oven, microwave, refrigerator, washing machine, car, safe, fridge, cabinet, window, lever door, round door, faucet, table, cart, and holistic (YCB) objects, which are sourced from the PartNet-Mobility~\cite{xiang2020partnetmobility}, UniDoorManip~\cite{li2024unidoormanip}, and YCB~\cite{calli2015ycb} datasets. The USD files for these objects are constructed by processing their original URDF files. The object grasp points are automatically located by detecting the object "handle" body in their URDF files. For those lid-structured objects that lack a "handle" body, such as boxes, laptops, toilets, trashcans, and washing machines, we manually label the grasp points and initialize the lid as opened by 10 degrees for the open skill. For Holistic objects, the center of mass is used as the grasp point.

\noindent\textbf{Scene Assets.}
MobileManiBench features 100 realistic scene placements for trajectory generation. It collects 6 scene assets from the Isaac Sim~\cite{NVIDIA_Isaac_Sim} and 4 scene assets from the Genie Sim~\cite{contributors2025geniesimrepo}, with each scene asset contains multiple rooms. MobileManiBench places the 20 categories of objects into 5 realistic scene settings for trajectory rendering: space (cart), wall (toilet, trashcan, refrigerator, washing machine, fridge, cabinet, table), door (window, lever door, round door), outdoor (car), and tabletop (box, laptop, dishwasher, oven, microwave, safe, faucet, holistic objects). For each scene setting, we manually annotate 20 scene placements within the scene assets, as shown in Figure~\ref{fig:space} -~\ref{fig:outdoor}.

\subsection{Environment Setup}

\noindent\textbf{Initialization.}
Each environment contains one robot and one object placed either on the ground or on a tabletop with random heights ranging from 0.2 to 0.7 meters. The object's grasp point initialized to zero x and y coordinates, as shown in Figure~\ref{fig:init}. The robot is first initialized with the same pose, randomly within the orange zone, facing the object grasp point, and then rotating around the z axis from ${-15}^{o}$ to ${15}^{o}$. For the open skill, the goal points for articulated objects are set to the positions of their grasp points when the object is 60\% open. For the close skill, the goal points are set as the positions of the grasp points when the object is closed, with the object initialized from 40\% to 80\% opened. For carts and holistic (YCB) objects, a target displacement and height of 0.2 meters are set for pull, push, and pick tasks.

\noindent\textbf{Skill Definition.}
MobileManiBench designs 5 mobile manipulation skills: \textit{open} and \textit{close} for objects including boxes, toilets, laptops, trashcans, dishwashers, ovens, microwaves, refrigerators, washing machines, cars, safes, fridges, cabinets, windows, lever doors, round doors, faucets, and tables; \textit{pull} and \textit{push} for carts; and \textit{pick} for holistic objects. Each robot–object–skill combination is parameterized using keypoint-based displacements of the robot's gripper/hand points, the object grasp point, and the goal point. A manipulation is considered successful if the object grasp point reaches the goal point within a threshold of 5cm.

\begin{table*}[!t]
    \centering
    \resizebox{0.99\textwidth}{!}{
    \begin{tabular}{p{4.5cm}|p{11.5cm}}
        \toprule
Input Type      & Elements (Dimension) \\
\midrule
Time (30) & Sine-cosine time embedding (30).\\
\midrule
Object State (9) & Object grasp point position (3) and rotation (3); Object goal point position (3).\\
\midrule
Robot Proprioception (78/135) & Robot palm position (3), rotation (3), linear velocity (3), and angular velocity (3); Robot fingertip positions (9/15), rotations (9/15), linear velocities (9/15), and angular velocities (9/15); Robot active joint angles (10/21), joint velocities (10/21) and joint accelerations (10/21). \\
\midrule
Robot-Object Distance (22/31) & Robot palm to object grasp point distance (1); Robot fingertips to object grasp point distances (3/5); Robot active joints to object grasp point distances (15/22); Robot palm to object grasp point displacement (3). \\
\midrule
Previous Action (7/18) & Previous target robot wrist position displacement (3) and rotation displacement (3); Previous target robot finger joint angles (1/12).  \\

\bottomrule
    \end{tabular}}
    \vspace{2mm}
    \caption{Input types of MobileManiRL for the G1 robot and the XHand robot.}
    \label{tab:rl_input_details}
    \vspace{-8mm}
\end{table*}

\subsection{MobileManipRL Training}

MobileManiBench trains a universal state-based MobileManiRL on each robot-object-skill pair. It utlizes PPO~\cite{PPO} as the optimization algorithm.

\noindent\textbf{Action Space.}
Each robot is controlled with (6+D) dimensional actions, where the first 6 dimensions represent the wrist pose displacement relative to the previous frame, from which inverse kinematics computes the target joint angles of mobile base (2-d) and right arm (7-d). The remaining D dimensions correspond to the target joint angles of end effector. This formulation yields 7-d actions for the G1 robot and 18-d actions for the XHand robot.

\noindent\textbf{Observation Space.}
Table~\ref{tab:rl_input_details} summarizes the 5 input types used in MobileManiRL. Specifically, \textit{Time} encodes the timestep embedding; \textit{Object State} represents the states of the object grasp point and goal point; \textit{Robot Proprioception} captures the robot’s body and joint states; \textit{Robot–Object Distance} measures the distances between the robot gripper/hand points and the object grasp point; and \textit{Previous Action} records the previous (6+D)-dimensional action.

\noindent\textbf{Reward Function.}
The reward function described in Eq.(1) of the main paper comprises five components: $R_d$, $R_a$, $R_g$, $R_m$, and $R_s$. These reward components are governed by a grasp flag $f_g$, which indicates whether the gripper/hand has successfully grasped the object grasp point.

The distance reward $R_d$ penalizes the average Chamfer Distance ($CD$) between the gripper/hand points $P_{hand}^{i}$ and the object grasp point $P_{grasp}$, promoting contact and encouraging a secure grasp. It additionally  penalizes the Chamfer Distance between the object grasp point $P_{obj}$ and the goal point $P_{goal}$:
\begin{equation}
\label{eq:R_d}
R_{d} = -\omega_d \left( \frac{1}{N}\sum_{i=1}^{N} CD(P_{hand}^{i}, P_{obj}) + CD(P_{goal}, P_{obj}) \right),
\end{equation}
where the reward weight $\omega_d$ is set as 1.0, and $N$ is the number of gripper/hand points, as shown in Figure 2 of the main paper.

The grasp flag $f_{g}$ is set to 1 when the average Chamfer Distance between the gripper/hand points and the object grasp point falls below a predefined threshold $\lambda_c=0.1$:
\begin{equation}
\label{eq:f_g}
f_g = \mathbbm{1}\!\left[ \frac{1}{N}\sum_{i=1}^{N} CD(P_{hand}^{i}, P_{obj}) < \lambda_g \right],
\end{equation}
where $\mathbbm{1}[\cdot]$ denotes the indicator function.

Before the grasp is established, the approach action reward $R_a$ penalizes deviations of the current action $a$ from a reference approach action $a_{ref_a}$:
\begin{equation}
\label{eq:R_a}
R_a = -\omega_a \, \| a - a_{ref_a} \|_2,
\end{equation}
where the reward weight $\omega_a$ is set as 0.2. The reference approach action $a_{ref_a}$ consists of: (1) the displacement from the gripper/hand palm point $P_{palm}$ to the object grasp point $P_{obj}$, (2) the displacement from the current palm rotation $R_{palm}$ to the initial palm rotation $R_{palm}^{init}$, and (3) the initial joint angles for the end-effector joints:
\begin{equation}
\label{eq:ref_a}
\begin{aligned}
a_{ref_a}[0:3] &= P_{obj} - P_{palm}, \\
a_{ref_a}[3:6] &= R_{palm}^{init} - R_{palm}, \\
a_{ref_a}[6:6+D] &= \text{Joint}^{init}.
\end{aligned}
\end{equation}
This reference approach action encourages the gripper/hand to stay open while approaching the object grasp point.

Once a grasp is established, the rewards $R_g$, $R_m$, and $R_s$ are introduced to guide the moving process. The grasp bonus $R_g$ encourages the grasp with $\omega_g=1.0$:
\begin{equation}
\label{eq:R_g}
R_g = \omega_g,
\end{equation}

The move action reward $R_{m}$ penalizes deviations of the current action $a$ from a reference move action $a_{ref_a}$:
\begin{equation}
\label{eq:R_m}
R_m = -\omega_m \, \| a[0:3] - a_{ref_m}[0:3] \|_2,
\end{equation}

\begin{equation}
\label{eq:ref_m}
\begin{aligned}
a_{ref_m}[0:3] &= P_{goal} - P_{palm}
\end{aligned}
\end{equation}
where $\omega_m$ is set as 0.2. This reference move action encourages the gripper/hand to move to the goal point.

The success reward $R_{s}$ provides a bonus when the object grasp point successfully reaches the goal point, defined by a threshold $\lambda_g=0.05$:
\begin{equation}
\label{eq:R_s}
R_s = \omega_s \mathbbm{1}[\parallel P_{goal} - P_{obj} \parallel_2 < \lambda_g],
\end{equation}
where $\omega_s$ is set as 2.0.

\begin{table*}[!t]
%\small
%\setlength\tabcolsep{5pt}
    \centering
    % \resizebox{\linewidth}{!}{
    \resizebox{0.99\textwidth}{!}{\begin{tabular}{l|ccc|cccc|cccc}
        \toprule
        & \multicolumn{3}{c|}{Number} & \multicolumn{4}{c|}{MobileManiRL (\%)} & \multicolumn{4}{c}{MobileManiVLA (\%)} \\
        \cmidrule(lr){2-4} \cmidrule(lr){5-8} \cmidrule(lr){9-12}
        Object & \multirow{2}{*}{Total} & \multirow{2}{*}{Train} & \multirow{2}{*}{Test} & \multicolumn{2}{c}{Open/Pull/Pick} & \multicolumn{2}{c|}{Close/Push} & \multicolumn{2}{c}{Open/Pull/Pick} & \multicolumn{2}{c}{Close/Push} \\
        \cmidrule(lr){5-8} \cmidrule(lr){9-12}
         & & & & G1 & XHand & G1 & XHand & G1 & XHand & G1 & XHand \\
        \midrule
        Box              & 14 & 10 & 4 & 71.2 & 91.4 & 81.2 & 92.0 & 55.0 & 22.5 & 61.9 & 95.0 \\
        Toilet           & 30 & 25 & 5 & 89.6 & 94.6 & 98.2 & 99.6 & 63.5 & 42.0 & 64.5 & 95.0 \\
        Laptop           & 36 & 30 & 6 & 91.2 & 98.4 & 98.5 & 88.9 & 74.6 & 64.6 & 67.9 & 98.8 \\
        Trashcan         & 36 & 30 & 6 & 88.7 & 93.9 & 95.8 & 99.8 & 41.4 & 37.9 & 51.8 & 76.4 \\
        Dishwasher       & 25 & 20 & 5 & 98.8 & 99.7 & 99.4 & 99.4 & 93.5 & 86.0 & 99.5 & 97.0 \\
        Oven             & 12 & 10 & 2 & 99.7 & 99.9 & 99.8 & 99.7 & 93.8 & 77.5 & 90.0 & 97.5 \\
        \midrule
        Microwave        &  5 &  3 & 2 & 97.4 & 95.0 & 99.2 & 99.6 & 32.5 & 42.5 & 42.2 & 85.0 \\
        Refrigerator     & 52 & 40 &12 & 72.9 & 73.2 & 99.1 & 99.8 & 18.1 & 12.8 & 91.3 & 77.5 \\
        Washing Machine  & 12 & 10 & 2 & 76.6 & 90.0 & 97.7 & 99.5 & 40.0 & 67.5 & 65.0 & 67.5 \\
        Car              & 12 & 10 & 2 & 80.3 & 97.9 & 98.4 & 97.0 & 28.7 & 26.2 & 97.5 & 90.0 \\
        Safe             & 19 & 15 & 4 & 81.8 & 83.1 & 98.3 & 97.4 & 35.6 & 30.0 & 96.9 & 98.8 \\
        Fridge           & 10 &  8 & 2 & 90.4 & 99.1 & 99.7 & 99.7 & 37.5 & 10.0 & 95.0 & 95.0 \\
        Cabinet          & 22 & 20 & 2 & 84.2 & 84.4 & 98.5 & 99.4 & 40.0 & 33.8 & 91.2 & 91.2 \\
        Window           & 18 & 15 & 3 & 88.4 & 93.4 & 99.1 & 99.2 & 34.2 & 11.7 & 90.8 & 94.2 \\
        Lever Door       & 22 & 18 & 4 & 90.0 & 85.8 & 95.2 & 95.6 & 55.6 & 58.1 & 94.0 & 94.5 \\
        Round Door       & 25 & 20 & 5 & 86.3 & 91.1 & 95.7 & 96.0 & 41.0 & 71.5 & 92.5 & 95.7 \\
        \midrule
        Faucet           & 78 & 60 &18 & 97.7 & 99.8 & 86.3 & 87.5 & 41.4 & 31.7 & 34.3 & 38.1 \\
        \midrule
        Table            &108 & 90 &18 & 81.0 & 92.2 & 99.9 & 99.9 & 25.8 & 19.7 & 96.4 & 93.9 \\
        Cart             & 16 & 12 & 4 & 80.8 & 97.3 & 93.1 & 97.2 & 34.4 & 29.4 & 85.0 & 90.0 \\
        \midrule
        Holistic (YCB)   & 78 & 60 &18 & 66.4 & 72.6 &  -   &  -   & 28.8 & 40.2 &  -   &  -   \\
        \midrule \midrule
       \textbf{}{Summary}&630 &506 &124& 83.9 & 89.7 & 96.2 & 96.5 & 40.3 & 36.5 & 75.8 & 81.5 \\
        \bottomrule
    \end{tabular}}
    \vspace{2mm}
    \caption{MobileManiBench object numbers and success rates of MobileManiRL and MobileManiVLA.}
    \label{tab:object_details}
    \vspace{-4mm}
\end{table*}

\begin{figure}[!t]
    \centering
    \includegraphics[width=0.8\linewidth]{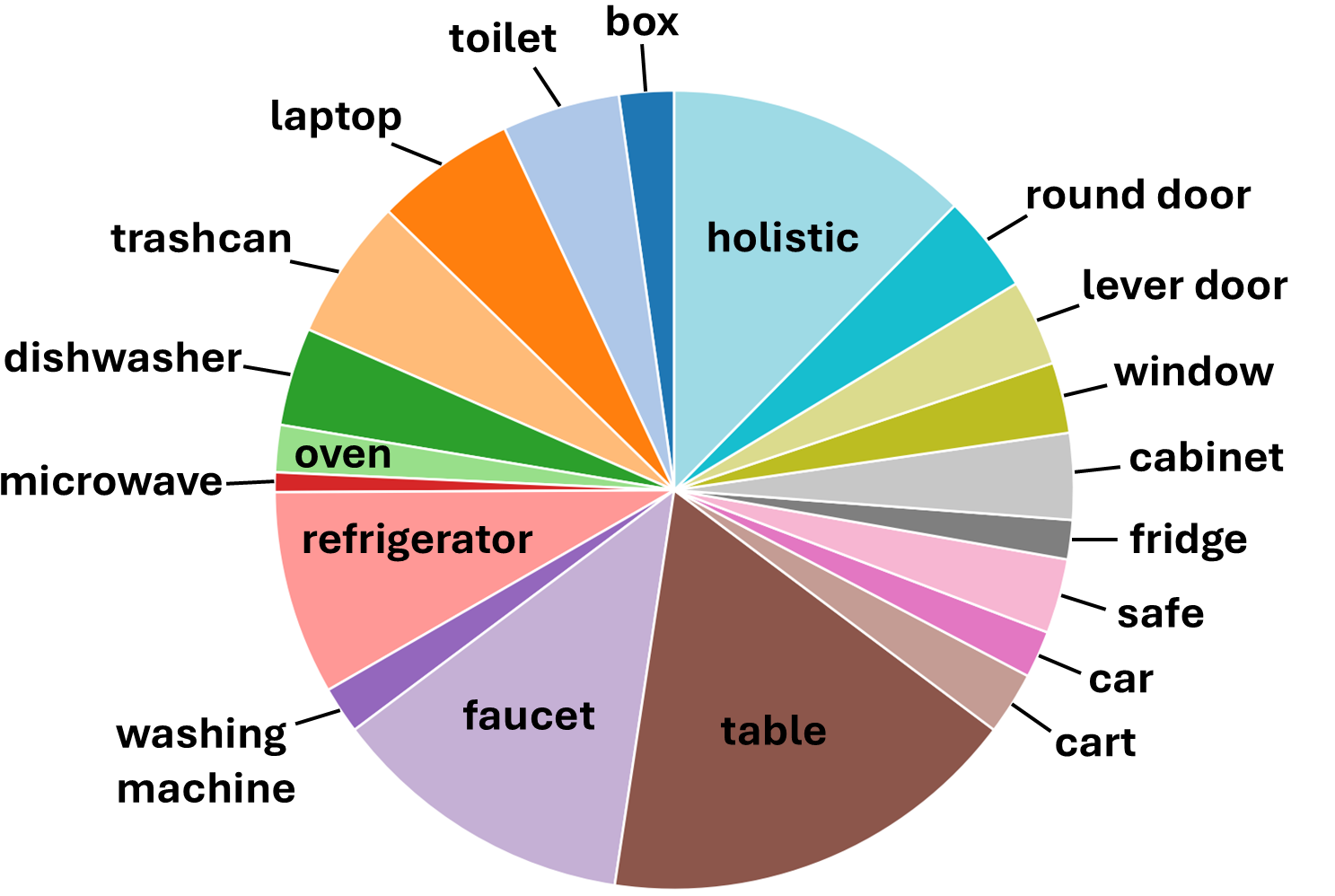}
    \vspace{2mm}
    \caption{
    MobileManiBench object distribution.
    }
    \vspace{-3mm}
    \label{fig:object_distribute}
\end{figure}

\subsection{MobileManiDataset Generation}
MobileManiBench generates the MobileManiDataset by rendering MobileManiRL in diverse realistic scenes.

\noindent\textbf{Realistic Scenes.}
We place the 20 categories of objects into 5 realistic scene settings for trajectory rendering: space, wall, door, tabletop, and outdoor. For each setting, we manually annotate 20 scene placements using digital assets from the Isaac Sim~\cite{NVIDIA_Isaac_Sim} and Genie Sim~\cite{contributors2025geniesimrepo}, yielding 80 for training and 20 for testing, as shown in Figure~\ref{fig:space} -~\ref{fig:outdoor}.

\noindent\textbf{Dataset Analysis.}
MobileManiDataset comprises a total of 630 objects across 20 categories, as shown in Table~\ref{tab:object_details} and Figure~\ref{fig:object_distribute}, which are manipulated through 5 skills and over 100 tasks using 1,182 robot-object-skill combinations of MobileManiRL. The dataset is split into 506 objects and 80 scenes for VLA training, plus 124 objects and 20 scenes for testing, yielding a total of 15,232 robot-object–skill–scene combinations for training and 920 combinations for testing. For each training combination, we generate 10 successful manipulation trajectories in Isaac Sim, resulting in 150K training trajectories. For each robot, the numbers of training trajectories for the open, close, pull, push, and pick skills are 69.4k, 69.4k, 1.9k, 1.9k, and 9.6k, respectively.
Each manipulation trajectory, $\mathcal{T}=\{L, (I_1, S_1,A_1),\dots, (I_t,S_t,A_t), \dots, (I_T,S_T,A_T)\}$, is recorded at 30 FPS with an average length of 160 frames, including one natural language instruction $L$ of the form “\textless skill\textgreater\ \textless object\textgreater”; synchronized $520\times520\times3$ RGB, depth, and segmentation images $I_t$ from both head-view and wrist-view cameras; the corresponding object and robot states $S_t$ in Table~\ref{tab:rl_input_details}; and the executed (6+D) dimensional action $A_t$ at each timestep $t$. All states and actions are recorded in the global world coordinate frame. Each depth image is obtained by first clipping the raw depth values to 0–5 meters and subsequently normalized to pixel values between 0 and 255. Manipulation trajectories for the G1 robot and the XHand robot are shown in Figure~\ref{fig:open} -~\ref{fig:pick}.

\subsection{MobileManiVLA}
Section 4.3 illustrates the model architecture of MobileManiVLA. Our framework incorporates multiple modalities as inputs, including multi-view RGB images, multi-view depth images, language instructions, and robot state information. The multi-view RGB and multi-view depth images are each passed through a single SigLIP encoder to obtain their respective embeddings. These embeddings are then projected into the same representation space as the language embeddings and  concatenated together, along with the language tokens and a cognition token. Then they are fed into the Gemma-2 language model.

The final-layer embedding of the cognition token produced by Gemma-2, together with the robot state, serves as the conditioning for the diffusion model. This conditioning, combined with a noisy action chunk, is processed by a DiT-Base model (about 90M parameters). Through a 10-step DDIM denoising procedure, the model iteratively generates clean delta end-effector poses and gripper/hand actions, from which inverse kinematics computes the target joint angles of mobile base and right arm.

\section{Experiment Details}
\label{sec:experiment_supp}

\subsection{Main Results}

For each of the G1 robot and XHand robot, we first train MobileManiRL on each of the 1,182 robot-object-skill combinations, which provide 15,232 robot-object-skill-scene combinations for VLA training and 920 combinations for VLA testing when generating the MobileManiDataset, yielding 150K training trajectories for MobileManiVLA training.
Table~\ref{tab:object_details} demonstrates the detailed success rates of MobileManiRL and MobileManiVLA across the 20 object categories. MobileManiRL is evaluated on seen objects with 1024 episodes per robot-object-skill combination, whereas MobileManiVLA is evaluated on unseen objects and scenes, with 10 episodes per robot-object-skill-scene combination. Both experiments are conducted with randomized robot initial poses to assess robustness.

\subsection{Ablation Studies.}
\vspace{-2mm}
We conduct ablation studies and model comparisons on a subset of MobileManiDataset that includes only the challenging skills: open, pull, and pick for the G1 robot, covering 272 training and 66 testing objects across 6 categories: laptop, cabinet, faucet, table, cart and holistic objects, providing 4,352 robot-object-skill-scene combinations for training and 264 unseen combinations for testing. All VLA models are trained with a batch size of 120 and 160K iterations, which are evaluated on unseen objects and scenes with 10 episodes per combination.
The detailed success rates of MobileManiVLA on the G1 robot with different image inputs, state inputs, and unseen items are shown in Table~\ref{tab:supp_ablation_image_inputs}, Table~\ref{tab:supp_ablation_state_inputs}, and Table~\ref{tab:supp_ablation_unseen_items}.

\vspace{-6mm}
\begin{table}[h]
%\small
%\setlength\tabcolsep{5pt}
    \centering
    
    \resizebox{0.6\linewidth}{!}{
    \begin{tabular}{l|cccc}
        \toprule
        \multirow{2}{*}{Object} & Multi-view & Multi-view & Head-view & Head-view \\
        & RGB-D & RGB & RGB-D & RGB \\
        \midrule
        Laptop      & 72.5 & 40.8 & 30.0 & 16.9 \\
        Cabinet     & 22.3 & 13.8 & 12.5 & 6.8 \\
        Faucet      & 30.6 & 12.0 & 12.7 & 8.3 \\
        Table       & 19.3 & 13.4 & 8.9 & 7.1 \\
        Cart        & 23.5 & 13.8 & 9.4 & 6.3 \\
        Holistic    & 21.7 & 11.2 & 16.7 & 5.9\\
        \midrule
        Mean        & 28.2 & 14.9 & 14.1 & 7.9 \\
        \bottomrule
    \end{tabular}}
    \vspace{2mm}
    \caption{Success rates of MobileManiVLA on G1 Robot with different image inputs.}
    \label{tab:supp_ablation_image_inputs}
    \vspace{-8mm}
\end{table}

\vspace{-10mm}
\begin{table}[h]
%\small
%\setlength\tabcolsep{5pt}
    \centering
    \resizebox{0.4\linewidth}{!}{
    \begin{tabular}{l|cccc}
        \toprule
        \multirow{2}{*}{Object} & \multirow{2}{*}{None} & Hand & + Grasp & + Goal \\
        &  & Pose & Point & Point \\
        \midrule
        Laptop      & 59.4 & 72.5 & 83.3 & 85.4 \\
        Cabinet     & 18.8 & 22.3 & 23.8 & 32.5 \\
        Faucet      & 27.1 & 30.6 & 36.4 & 40.4 \\
        Table       & 12.5 & 19.3 & 20.7 & 24.8 \\
        Cart        & 28.1 & 23.5 & 25.0 & 32.8 \\
        Holistic    & 14.5 & 21.7 & 25.5 & 29.5 \\
        \midrule
        Mean        & 22.4 & 28.2 & 32.3 & 36.6 \\
        \bottomrule
    \end{tabular}}
    \vspace{2mm}
    \caption{Success rates of MobileManiVLA on the G1 Robot with different state inputs.}
    \label{tab:supp_ablation_state_inputs}
    \vspace{-8mm}
\end{table}

\vspace{-10mm}
\begin{table}[!h]
%\small
%\setlength\tabcolsep{5pt}
    \centering
    \resizebox{0.7\linewidth}{!}{
    \begin{tabular}{l|cccc}
        \toprule
        \multirow{2}{*}{Object} & Unseen Object & Unseen Object & Seen Object & Seen Object \\
        & Unseen Scene & Seen Scene & Unseen Scene & Seen Scene \\
        \midrule
        Laptop      & 72.5 & 90.0 & 73.8 & 87.5 \\
        Cabinet     & 22.3 & 30.0 & 56.8 & 65.7 \\
        Faucet      & 30.6 & 39.9 & 52.5 & 59.0 \\
        Table       & 19.3 & 25.3 & 57.5 & 61.2 \\
        Cart        & 23.5 & 33.1 & 44.5 & 50.5 \\
        Holistic    & 21.7 & 37.7 & 37.2 & 50.5 \\
        \midrule
        Mean        & 28.2 & 39.2 & 51.3 & 59.6 \\
        \bottomrule
    \end{tabular}}
    \vspace{2mm}
    \caption{Success rates of MobileManiVLA on the G1 Robot with seen or unseen objects and scenes.}
    \label{tab:supp_ablation_unseen_items}
    \vspace{-4mm}
\end{table}

\begin{table}[h]
%\small
%\setlength\tabcolsep{5pt}
    \centering
    \resizebox{0.6\linewidth}{!}{
    \begin{tabular}{l|ccccc}
        \toprule
        Object      & MobileManiVLA & $\pi_{0.5}$ & $\pi_{0}$ & CogACT & OpenVLA \\
        \midrule
        Laptop      & 72.5 & 47.9 & 16.3 & 13.7 & 8.4 \\
        Cabinet     & 22.3 & 20.0 & 23.7 & 13.8 & 3.7 \\
        Faucet      & 30.6 & 23.3 & 16.5 & 6.0  & 4.3 \\
        Table       & 19.3 & 14.3 & 10.6 & 6.1  & 4.8 \\
        Cart        & 23.5 & 16.9 & 5.6  & 11.9 & 3.1 \\
        Holistic    & 21.7 & 9.7  & 4.5  & 4.2  & 3.5 \\
        \midrule
        Mean        & 28.2 & 18.8 & 11.2 & 6.8  & 4.5 \\
        \bottomrule
    \end{tabular}}
    \vspace{2mm}
    \caption{Model comparisons of representative VLA models with their default architectures and input modalities on the G1 robot.}
    \label{tab:supp_model_comparisons}
    \vspace{-8mm}
\end{table}

\subsection{Model Comparisons}
For the VLA model comparisons in Table 6 of the main paper, we fine-tune all models using a unified action representation. Specifically, each model predicts a 6D end-effector pose along with a gripper open/close action. The batch size is fixed at 120, and each model is trained for 160K steps. The learning schedule for each model is set according to its recommended or default configuration.

Specifically, $\pi_0$ was optimized using a cosine decay schedule with a 1,000 step warmup, a peak learning rate of 2.5e-5, decay steps of 100K, and a final decay learning rate of 2.5e-6. The action chunk size was 50 steps, and execution was performed over 8 steps. $\pi_{0.5}$ used a cosine decay schedule with a 1,000 step warmup, a peak learning rate of 5e-5, decay steps of 100K, and a final decay learning rate of 5e-5, with an action chunk size of 16 steps and executes 8 steps. Both $\pi_0$ and $\pi_{0.5}$ took two RGB images (head-view and wrist-view) as input. $CogACT$ emploies a constant learning rate of 2e-5 with an action chunk size of 16 steps. $OpenVLA$ predicted the next-step 7-dimensional action with a constant learning rate of 2e-5. Both vallina $CogACT$ and $OpenVLA$ use a single head-view RGB image as input and did not take the robot state as input.

\vspace{-2mm}
\section{Real-world VLA Inferences}
\label{sec:real_world_supp}
\vspace{-2mm}

\subsection{Hardware Setup.}
\begin{figure}[h]
    \centering
    \includegraphics[width=0.7\linewidth]{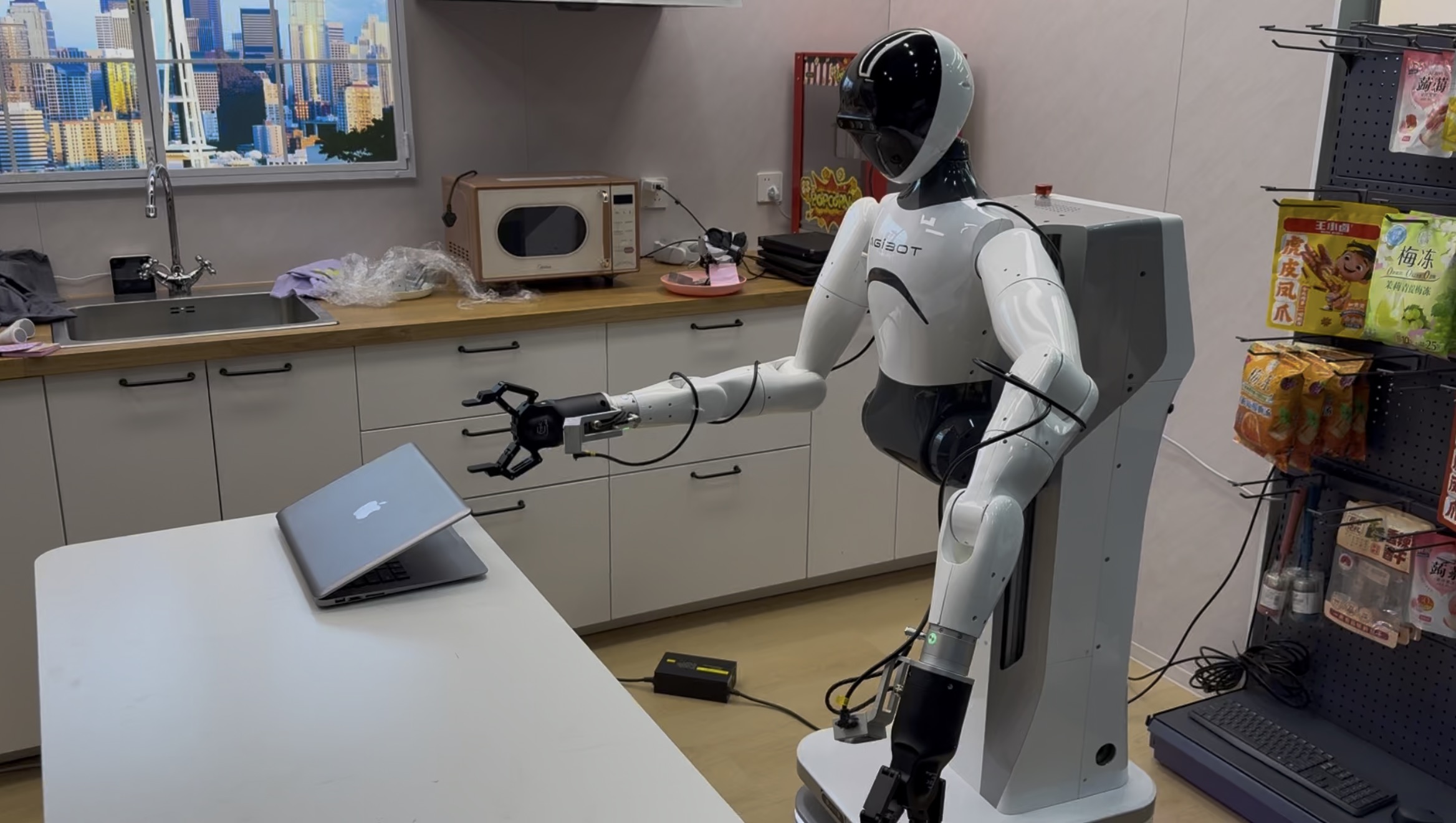}
    \vspace{-1mm}
    \caption{
    Real-world setup of the G1 robot.
    }
    \vspace{-2mm}
    \label{fig:real_world_g1}
\end{figure}

We inference our MobileManiVLA in the real-world on the G1 robot from AgiBot~\cite{agibot}, which is equipped with dual-arms, a head-mounted camera capturing images at $1280\times720$, and two wrist-mounted cameras capturing images at $640\times480$, as shown in Figure~\ref{fig:real_world_g1}. To align with our simulation setting, we fix the left arm and only active the mobile base, right arm, and right gripper. All raw head-view and wrist-view RGB images are cropped to $480\times480$ before being fed into the model. 

Due to hardware limitations, AgiBot does not provide access to the wrist cameras’ depth sensors, and the depth quality of the head camera is severely degraded. Therefore, for visual reliability, we adopt the MobileManipVLA variant that uses only the multi-view RGB images and hand pose state inputs, as reported in Table~\ref{tab:supp_ablation_image_inputs}. This RGB only MobileManipVLA achieves a 40.8\% success rate on the open laptop task but falls below 15\% on all other tasks. To ensure safety, our real-world inferences focus exclusively on the open laptop task.

\begin{figure*}[h]
    \centering
    \includegraphics[width=0.99\linewidth]{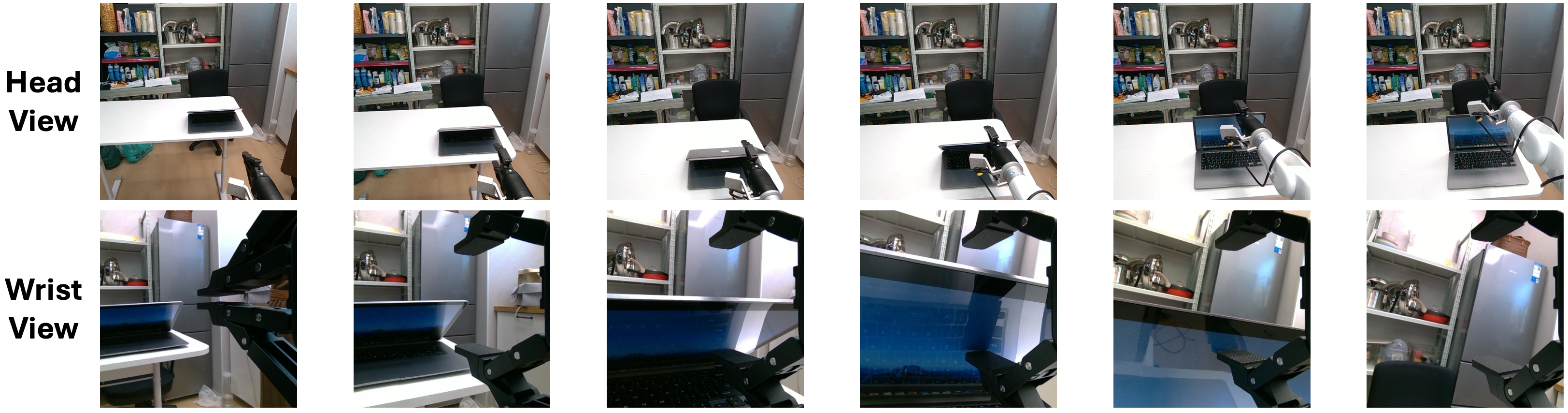}
    \vspace{-1mm}
    \caption{
    Real-world inference of MobileManiVLA on G1 robot for the open laptop task.
    }
    \vspace{-4mm}
    \label{fig:real_world_open_laptop}
\end{figure*}

\subsection{VLA Inference Results.}

For the open laptop task, we inference our MobileManiVLA with the robot base starts 1m away from the laptop. We inference the model over 10 trials with randomized robot pose initialization.
During inference, MobileManiVLA receives the natural-language instruction “open laptop,” real-time RGB images from the head and wrist cameras, and the real-time right-wrist pose expressed in the mobile base frame. Demonstrations of the RGB images usded during inference are shown in Figure~\ref{fig:real_world_open_laptop}. The overall success rate achieved in the real world is 40\% for the open laptop task.

\vspace{-4mm}

\bibliographystyle{splncs04}
\bibliography{main}

\begin{figure*}[h]
    \centering
    \includegraphics[width=0.99\linewidth]{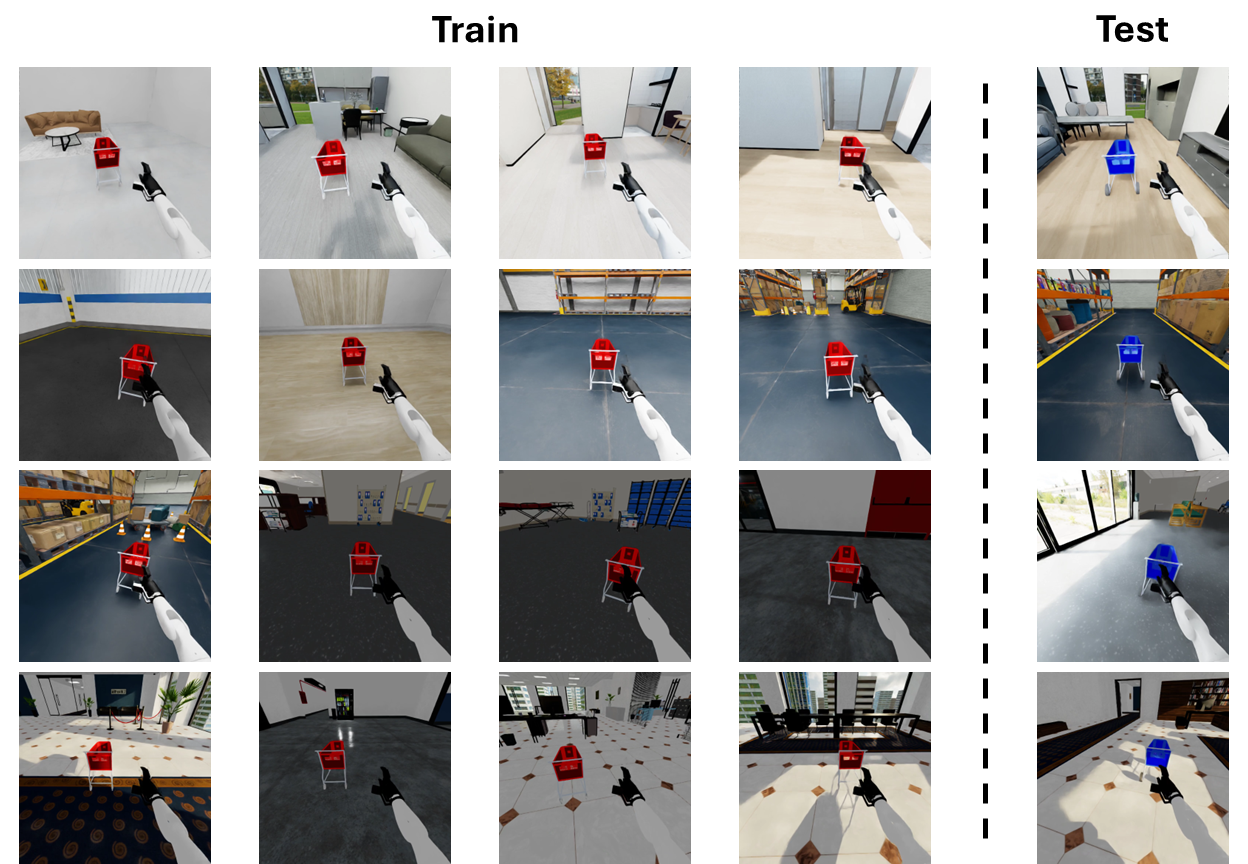}
    \vspace{-1mm}
    \caption{
    Realistic \textbf{space} scenes for cart.
    }
    \vspace{-2mm}
    \label{fig:space}
\end{figure*}

\begin{figure*}[h]
    \centering
    \includegraphics[width=0.99\linewidth]{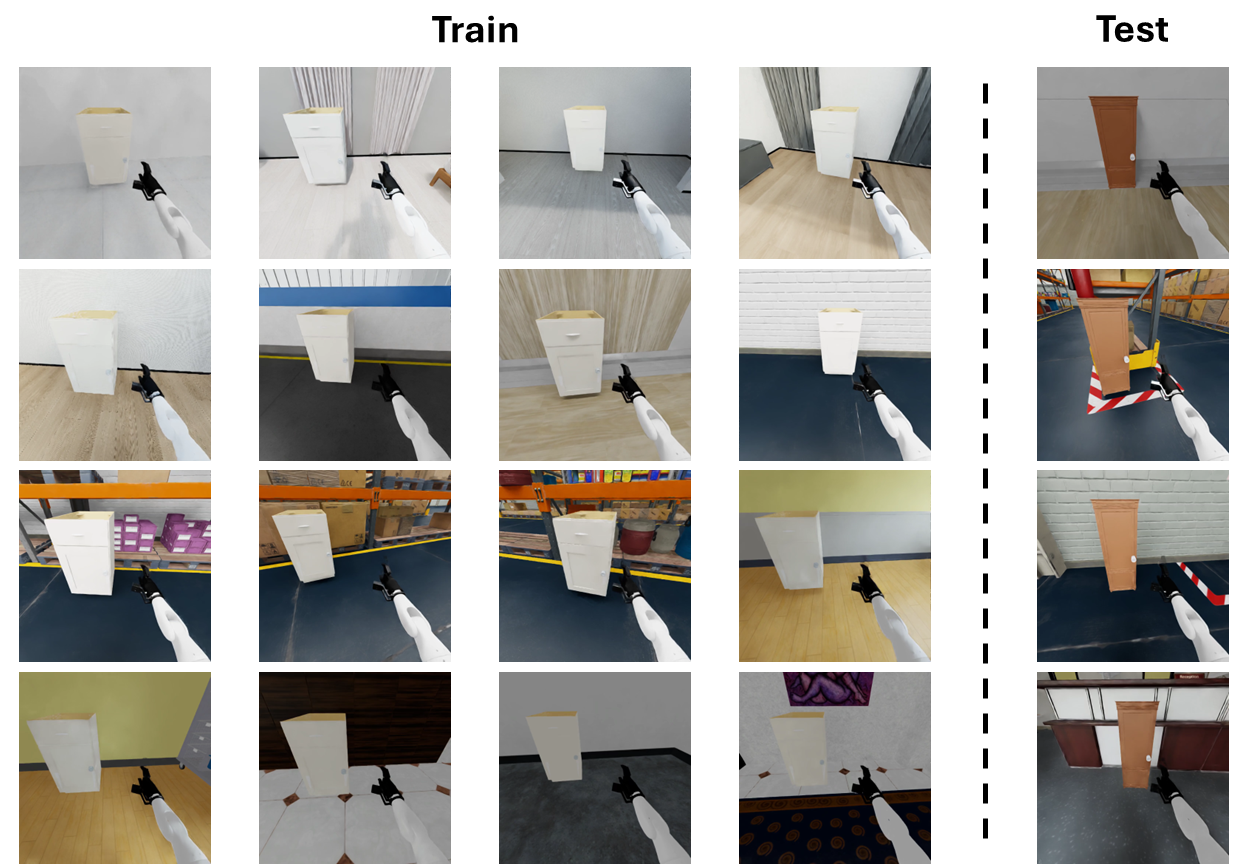}
    \vspace{-1mm}
    \caption{
        Realistic \textbf{wall} scenes for toilet, trashcan, refrigerator, washing machine, fridge, cabinet, table.
    }
    \vspace{-2mm}
    \label{fig:wall}
\end{figure*}

\begin{figure*}[h]
    \centering
    \includegraphics[width=0.99\linewidth]{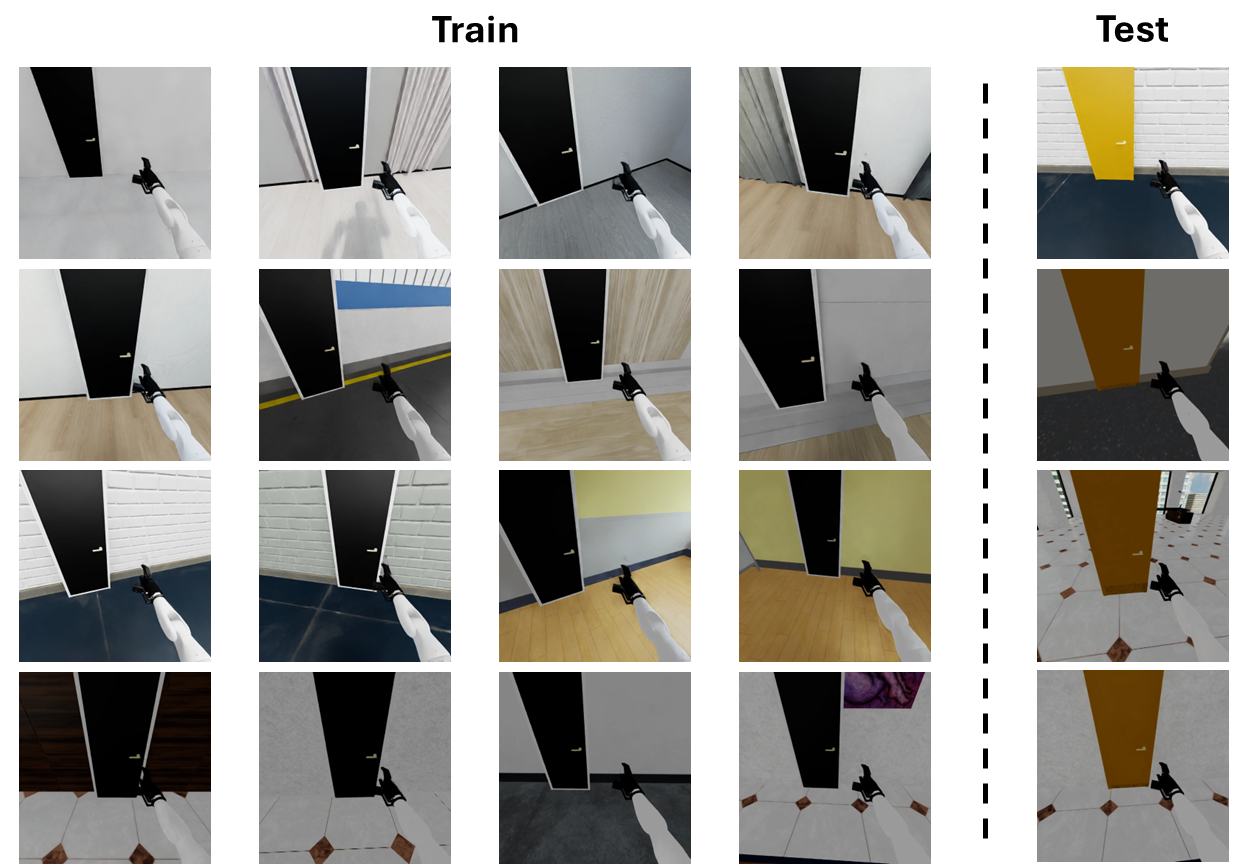}
    \vspace{-1mm}
    \caption{
        Realistic \textbf{door} scenes for window, lever door, round door.
    }
    \vspace{-2mm}
    \label{fig:door}
\end{figure*}

\begin{figure*}[h]
    \centering
    \includegraphics[width=0.99\linewidth]{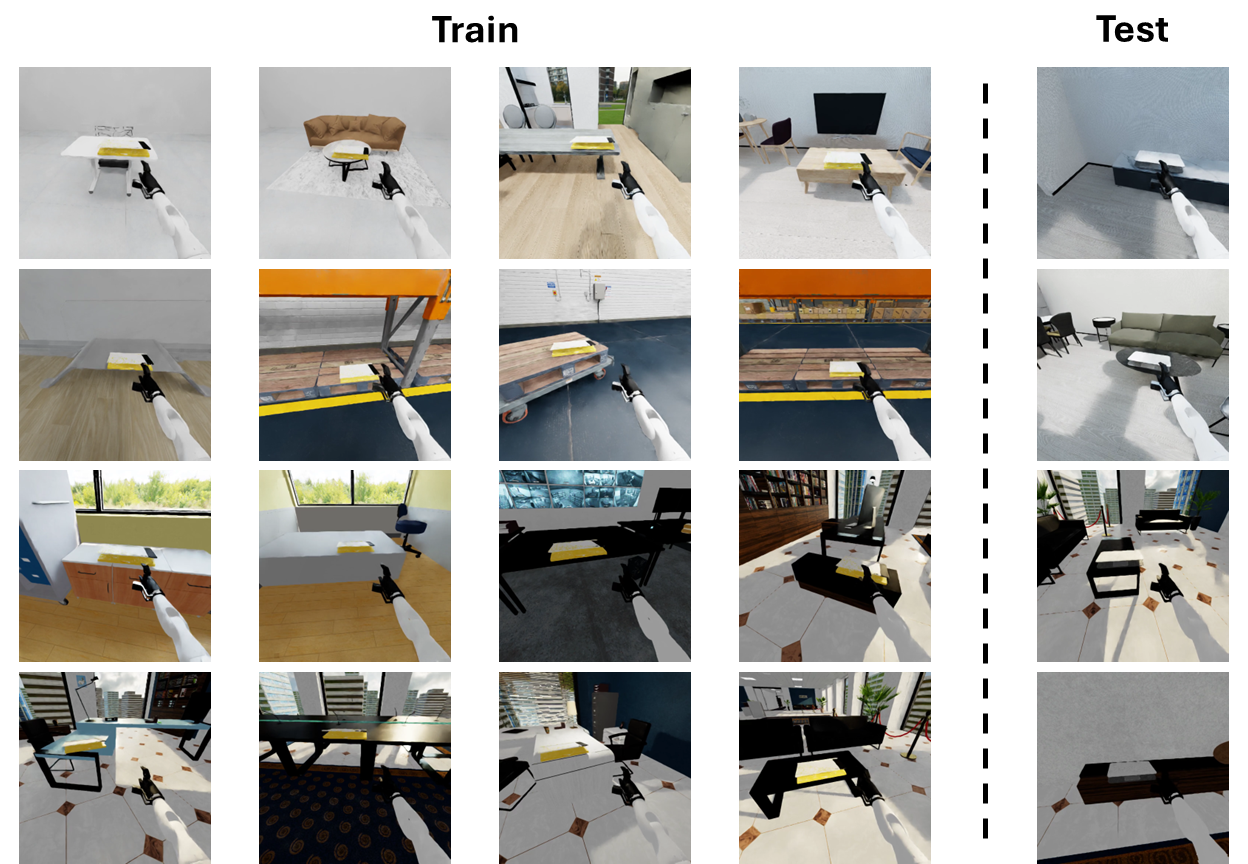}
    \vspace{-1mm}
    \caption{
        Realistic \textbf{tabletop} scenes for box, laptop, dishwasher, oven, microwave, safe, faucet, holistic objects.
    }
    \vspace{-2mm}
    \label{fig:tabletop}
\end{figure*}

\begin{figure*}[h]
    \centering
    \includegraphics[width=0.99\linewidth]{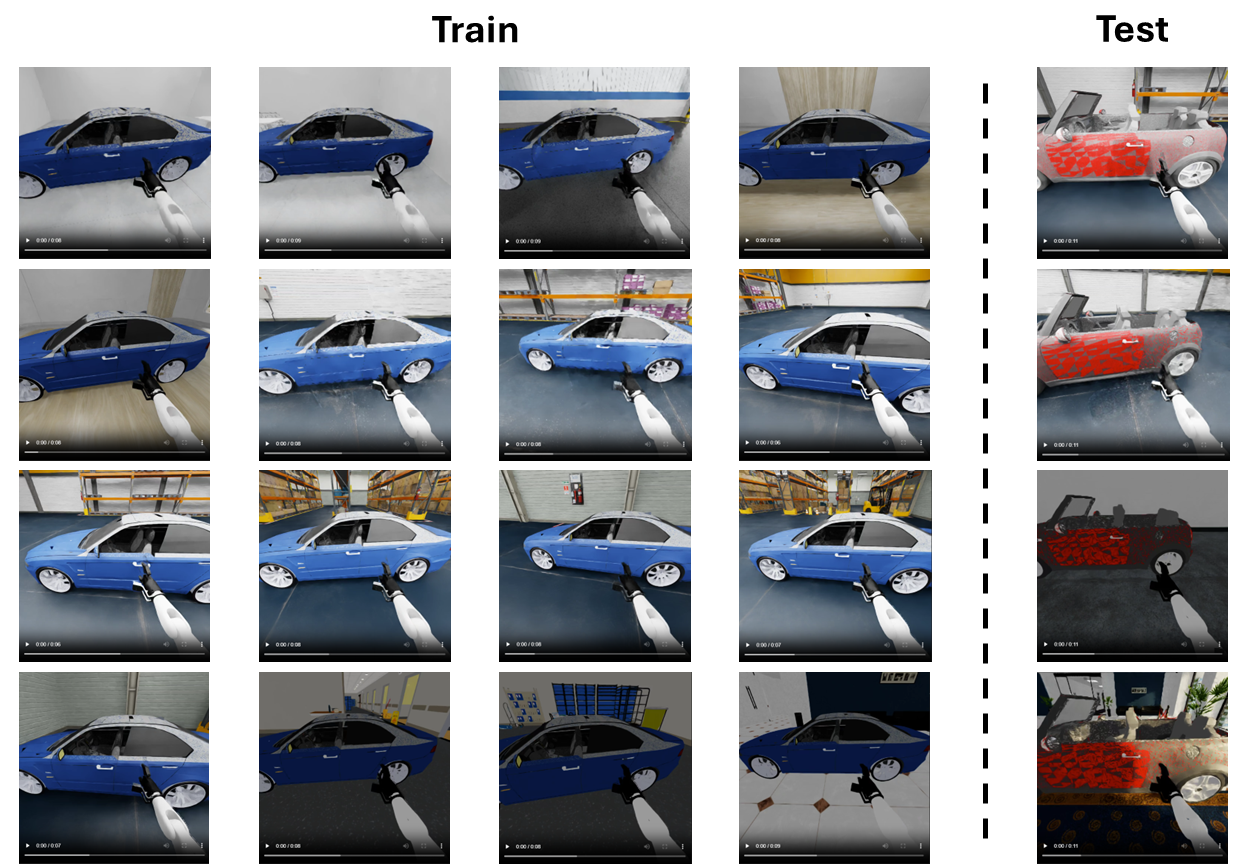}
    \vspace{-1mm}
    \caption{
        Realistic \textbf{outdoor} scenes for car.
    }
    \vspace{-2mm}
    \label{fig:outdoor}
\end{figure*}

\begin{figure*}[h]
    \centering
    \includegraphics[width=0.99\linewidth]{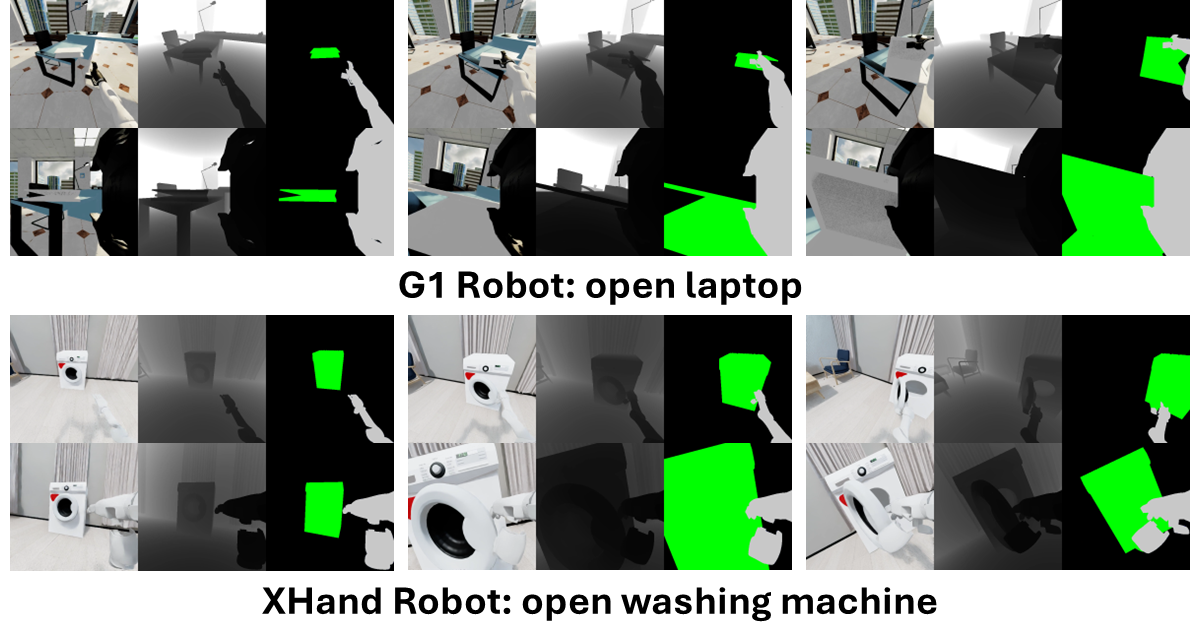}
    \vspace{-1mm}
    \caption{
    \textbf{Open} trajectories from the G1 robot and the XHand robot.
    }
    \vspace{-2mm}
    \label{fig:open}
\end{figure*}

\begin{figure*}[h]
    \centering
    \includegraphics[width=0.99\linewidth]{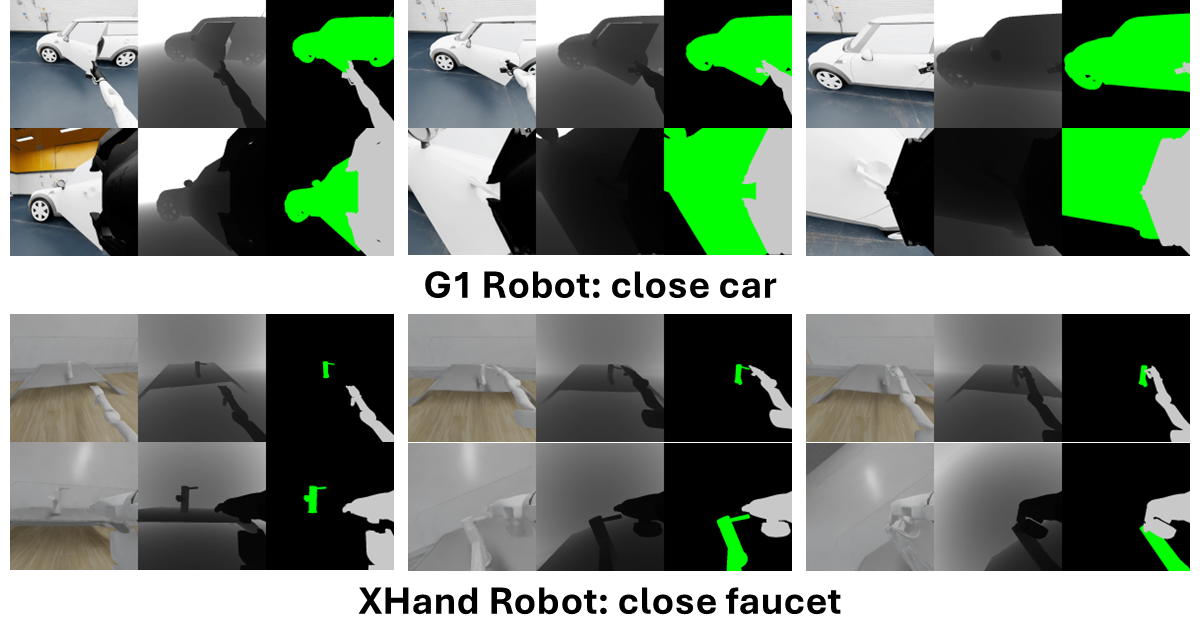}
    \vspace{-1mm}
    \caption{
    \textbf{Close} trajectories from the G1 robot and the XHand robot.
    }
    \vspace{-2mm}
    \label{fig:close}
\end{figure*}

\begin{figure*}[h]
    \centering
    \includegraphics[width=0.99\linewidth]{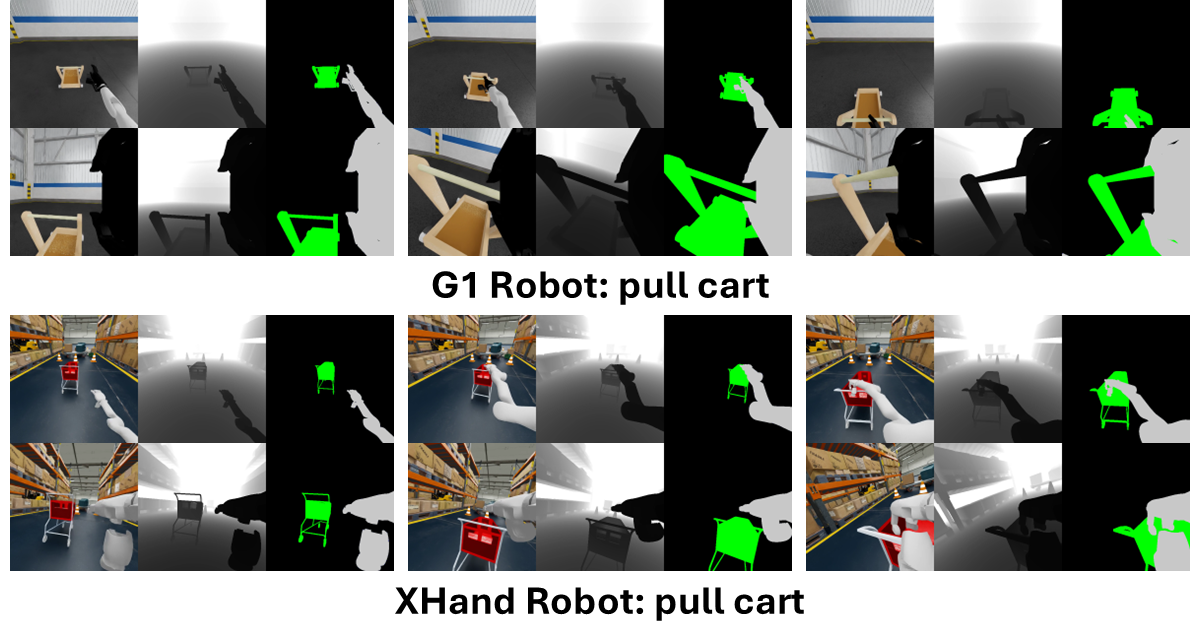}
    \vspace{-1mm}
    \caption{
    \textbf{Pull} trajectories from the G1 robot and the XHand robot.
    }
    \vspace{-2mm}
    \label{fig:pull}
\end{figure*}

\begin{figure*}[h]
    \centering
    \includegraphics[width=0.99\linewidth]{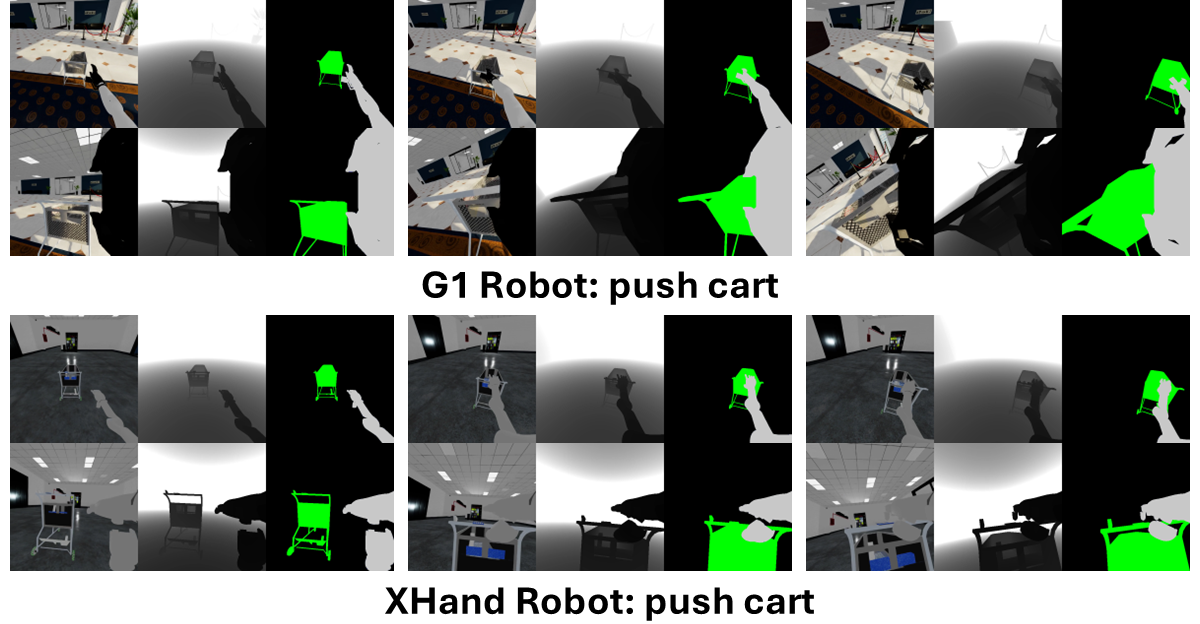}
    \vspace{-1mm}
    \caption{
    \textbf{Push} trajectories from the G1 robot and the XHand robot.
    }
    \vspace{-2mm}
    \label{fig:push}
\end{figure*}

\begin{figure*}[h]
    \centering
    \includegraphics[width=0.99\linewidth]{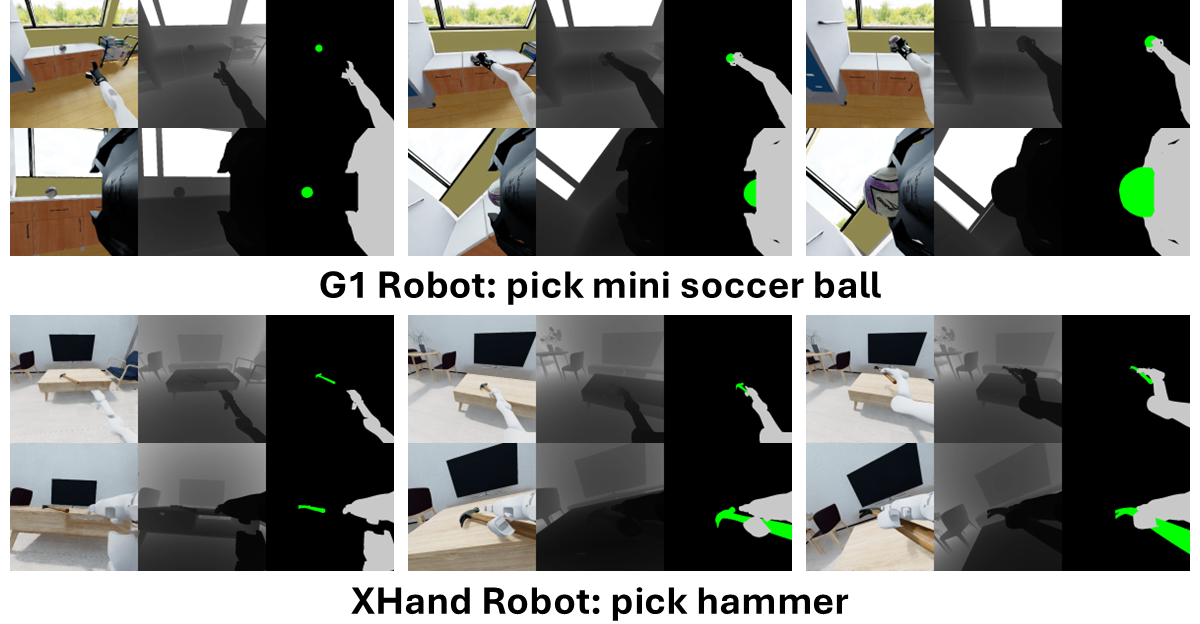}
    \vspace{-1mm}
    \caption{
    \textbf{Pick} trajectories from the G1 robot and the XHand robot.
    }
    \vspace{-2mm}
    \label{fig:pick}
\end{figure*}